\documentclass{article}

\PassOptionsToPackage{numbers,compress}{natbib}

\usepackage[preprint]{neurips_2026}

\usepackage[utf8]{inputenc} 
\usepackage[T1]{fontenc}    
\usepackage{hyperref}       
\usepackage{url}            
\usepackage{booktabs}       
\usepackage{amsfonts}       
\usepackage{nicefrac}       
\usepackage{microtype}      
\usepackage{amsmath}        
\usepackage{amssymb}        
\usepackage{algorithm}      
\usepackage{algorithmic}    
\usepackage{multirow}       
\usepackage{graphicx}       
\usepackage{subcaption}     
\usepackage{amsthm}

\usepackage[most]{tcolorbox}
\usepackage{graphicx}
\usepackage{tabularx}
\usepackage[useregional]{datetime2}
\DTMsetdatestyle{iso}  
\usepackage[normalem]{ulem}
\usepackage{wrapfig}
\usepackage{titlesec}
\usepackage{adjustbox}
\usepackage{enumitem}
\usepackage{booktabs}
\usepackage{multirow}
\usepackage{array}
\usepackage{float}
\usepackage{caption}
\usepackage{amsfonts}       
\usepackage{amsmath, amssymb}       
\usepackage{nicefrac}       
\usepackage{hyperref}       
\usepackage{url}            
\usepackage[utf8]{inputenc} 
\usepackage[T1,T5]{fontenc}    
\usepackage{natbib}
\usepackage{graphicx}
\usepackage{enumitem}
\usepackage{makecell}
\definecolor{tiiPurple}{RGB}{122, 0, 255}
\usepackage[x11names,dvipsnames,table]{xcolor}
\definecolor{bestcolor}{RGB}{220,255,220}
\usepackage{CJKutf8}
\usepackage[para]{threeparttable}

\usepackage{xspace}

\title{Falcon Perception-HD: High Density Perception via Reinforcement Learning}

\author{%
  Sofian Chaybouti\textsuperscript{1,2} \quad
  Yasser Dahou\textsuperscript{1} \quad
  Ngoc Dung Huynh\textsuperscript{1} \quad
  Reda Alami\textsuperscript{1} \quad
  Hilde Kuehne\textsuperscript{2,3} \\[0.4em]
  \textsuperscript{1}Technology Innovation Institute, Abu Dhabi, UAE \\
  \textsuperscript{2}T\"ubingen AI Center / University of T\"ubingen \qquad
  \textsuperscript{3}MIT-IBM Watson AI Lab \\
}

\begin{document}

\maketitle

\begin{abstract}

Autoregressive perception models trained to localize visual entities under the open-vocabulary setting are mostly trained using Supervised fine-tuning (SFT) with maximum likelihood, yet it optimizes a proxy objective (per-token cross-entropy) that is fundamentally misaligned with perception metrics such as precision and recall. In this paper, we explore post-training reinforcement learning (RL), specifically GRPO, to directly align these models with their evaluation metrics. Building up on the recently introduced Falcon Perception, we design an RL framework that addresses perception-specific challenges: reward design for set-structured outputs and multi-head sampling control. We discover multiple benefits from RL for perception: first, RL unlocks state-of-the-art performance in very dense scenes (up to 500 objects per scene), a regime where most existing systems degrade sharply or collapse; furthermore it fixes common issues in autoregressive perception models like mask repetitions and removes almost entirely the need for NMS and coordinate deduplication, which improve both performance and efficiency and remove the need for hyperparameters tuning; overall, we notice improvements on all levels of difficulties in referring expression segmentation (on PBench and SACO-Gold), and we find an elegant way to preserve the knowledge of whether an object exists or not (as evaluated by MCC) without training on negative samples. We show that a simple reward that penalizes false negatives and positives is sufficient. We develop two hybrid self-annotation pipelines, respectively tailored for difficult referring expressions and very dense scenes, and show their benefits on RL-training. Model weights are released as a Falcon Perception revision~\footnote{\url{https://huggingface.co/tiiuae/Falcon-Perception}}. Datasets will be published.

\end{abstract}

\section{Introduction}
\label{sec:intro}

\begin{figure}[t]
    \centering
    \includegraphics[page=1, width=\textwidth, trim={0 0 0 0}, clip]{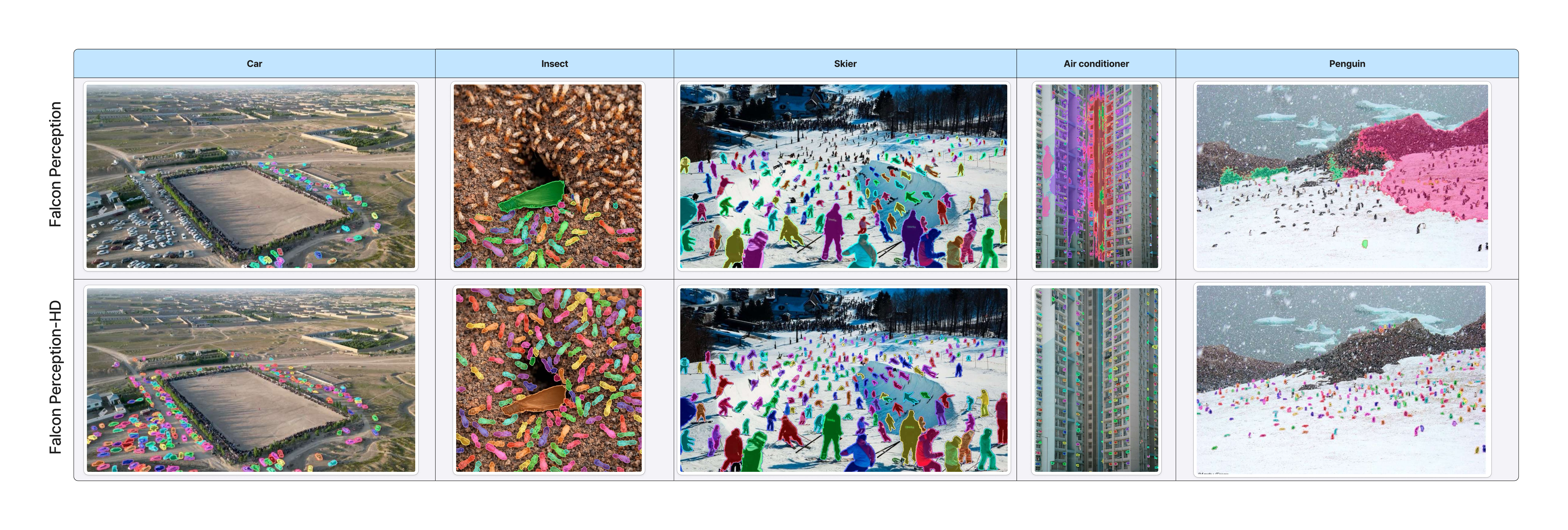}
    \caption{\textbf{Falcon Perception-HD on very dense scenes.} The most impressive effect of RL on top of Falcon Perception is the ability to solve high-density perception at a scale where prior systems degrade or collapse. Top: predictions of the SFT-only baseline (Falcon Perception). Bottom: predictions of Falcon Perception-HD after RL post-training. The base model suffers from low recall and collapse; RL fixes both.}
    \label{fig:teaser}
    \vspace{-1.5em}
\end{figure}

Reinforcement learning (RL) post-training has become indispensable for large language models (LLMs), aligning outputs with human preferences~\cite{ouyang2022instructgpt} and unlocking advanced reasoning via test-time scaling~\cite{guo2025deepseek}. Yet, its application to computer vision, particularly \emph{dense perception} (detecting, localizing, and segmenting objects), remains largely unexplored. 

Simultaneously, autoregressive perception models~\cite{pix2seq, bevli2026falcon} have gained traction, offering the theoretical capability to emit an unbounded number of instances. However, these models are traditionally trained via maximum likelihood estimation (MLE), optimizing per-token cross-entropy with respect to a fixed-order reference sequence (e.g., a raster scan). This creates a fundamental misalignment: perception is evaluated on order-invariant \emph{set-matching metrics} (precision, recall, and F$_1$). MLE optimizes a proxy~\cite{pinto2023tuning} and restricts the policy to ground-truth trajectories, leaving the model blind to the consequences of its own errors. Consequently, MLE-trained models often assign high likelihoods to degenerate sequences, leading to duplicated predictions, overlapping masks, and catastrophic sequence collapse in dense scenes~\cite{pinto2023tuning}.

In contrast, RL algorithms like GRPO~\cite{shao2024deepseekmath, liu2025drgrpo, wang2025gtpo} enable models to explore their own predictions and directly optimize set-matching rewards, raising a central question: \emph{can RL bridge the gap between autoregressive generation and spatial evaluation?} To investigate whether RL can overcome MLE failure modes such as sequence collapse or the reliance on post-hoc heuristics like NMS, we adopt Falcon Perception~\cite{bevli2026falcon}, an early-fusion perception model~\cite{chaybouti2025siglino}, as our base model. Its autoregressive ``chain-of-perception'' architecture is well-suited for this study: it supports unbounded instance generation in extreme-density scenarios (unlike constrained query-based models such as SAM~3~\cite{carion2025sam}) and generates high-resolution masks via a parallel head, avoiding the severe rollout latency of fully tokenizing dense outputs~\cite{pix2seq}. Operating with a compact 0.6B parameter count and a robust SFT baseline, Falcon Perception provides an ideal, tractable testbed. Leveraging this hybrid design, we sample from the discrete language, center, and size heads during training while holding the continuous segmentation head fixed, thereby cleanly isolating the impact of RL on visual perception.

Our investigation uncovers several non-trivial dynamics of RL post-training for visual perception. First, we observe a powerful cascading optimization effect: restricting RL action sampling exclusively to the discrete coordinate head is sufficient to drive holistic improvements in bounding box and mask quality. Because the chain-of-perception explicitly conditions subsequent heads on prior spatial predictions, learning to accurately point to object centers naturally forces the deterministic downstream heads to yield higher-quality masks. This effectively bypasses the need to sample a complex continuous action space. Second, the post-trained model structurally internalizes spatial suppression. It inherently learns to avoid duplicate coordinates and mask repetitions, effectively eliminating the need for post hoc heuristics such as NMS and center-deduplication thresholds. Removing these fragile hyperparameters not only accelerates inference but also resolves fundamental NMS failure modes, such as the incorrect suppression of depth-distributed or heavily overlapping objects.

Most notably, RL unlocks robust autoregressive generation in extreme-density regimes (100–600 instances per image). While standard MLE-trained models (including the strong Falcon Perception baseline) suffer from sequence collapse or severe recall degradation in crowded scenes, RL fundamentally stabilizes long-horizon spatial generation. By explicitly penalizing early termination and rewarding sustained recall, our post-training framework prevents sequence collapse and establishes state-of-the-art performance on dense benchmarks (e.g., the PBench dense split) using only 2.5k training samples.

\vspace{-1em}
\section{Related Work}
\label{sec:related}

\vspace{-1em}

\paragraph{Autoregressive perception.}
Casting perception as conditional sequence generation was pioneered by
Pix2Seq~\citep{pix2seq} and scaled in
unified models such as Unified-IO~\citep{lu2023unifiedio,lu2024unifiedio2},
OFA~\citep{wang2022ofa}, and Florence-2~\citep{xiao2024florence2}.
Multimodal LLMs subsequently expose detection and segmentation through grounding tokens, including LISA~\citep{lai2024lisa},
GLaMM~\citep{rasheed2024glamm}, PixelLM~\citep{ren2024pixellm},
VisionLLM-v2~\citep{wu2024visionllmv2}, the Qwen-VL
family~\citep{bai2023qwenvl,wang2024qwen2vl,qwen2025qwen3vl}, and
Moondream~\citep{korrapati2024moondream}. In parallel, query-based
open-vocabulary detectors~\citep{liu2024gdino,minderer2024owlv2,ren2024dinox,shen2024ape} and the Segment Anything family~\citep{kirillov2023sam,ravi2024sam2,carion2025sam}
deliver strong results but do not have a sampling capability. Falcon Perception~\citep{bevli2026falcon},
on which we build, is natively multimodal with a chain-of-perception
factorization (centers, sizes, masks) and emits an unbounded number of objects. All these models, however, are trained with maximum likelihood on a fixed-order target sequence; we are, to our knowledge, the first to apply RL post-training directly to such a backbone and to show that the resulting output-format issues (mask repetition, collapse on dense scenes, dependence on
NMS) are addressable with a tiny dataset compared to the pretraining and SFT corpus.

\vspace{-1em}

\paragraph{RL post-training for language models.}
RLHF on a learned preference model~\citep{ouyang2022instructgpt,bai2022rlhf}
optimized with PPO~\citep{schulman2017ppo}, and later DPO~\citep{rafailov2023dpo},
established the main LLM post-training recipes. The shift to verifiable rewards (RLVR)~\citep{lambert2024tulu3} was popularized by DeepSeek-R1~\citep{guo2025deepseek} via GRPO~\citep{shao2024deepseekmath}, spawning a wave of refinements that target entropy collapse, length bias, and credit assignment, including DAPO~\citep{yu2025dapo},
CISPO~\citep{minimax2025cispo}, Dr.GRPO~\citep{liu2025drgrpo}, and
REINFORCE++~\citep{hu2025reinforcepp}. We adopt GRPO with the dual-clip stabilization of VeRL~\citep{sheng2024hybridflow}. However, our setting differs fundamentally from LLM post-training in two critical dimensions. First, the reward is a spatial order-invariant set-matching function (IoU and precision/recall) evaluated over continuous outputs. Second, we must apply this machinery to a heterogeneous, multi-head action space rather than a homogeneous text policy. 

\vspace{-1em}

\paragraph{RL for vision language models.}
RL has only recently been applied to vision-language tasks.
Visual-RFT~\citep{liu2025visualrft} and Vision-R1~\citep{huang2025visionr1}
use GRPO with rule-based rewards (IoU, accuracy) for grounding and VQA;
R1-V~\citep{chen2025r1v}, LMM-R1~\citep{peng2025lmmr1}, and
VisionReasoner~\citep{liu2025visionreasoner} extend the recipe to
multimodal reasoning, and Seg-Zero~\citep{liu2025segzero} targets referring
expression segmentation. Closest to our setting, the concurrent Rex-Omni~\citep{jiang2026detect} casts detection as next-point prediction in a 3B VLM and reaches a similar finding: teacher forcing produces duplicate predictions, which they mitigate with a GRPO stage using geometry-aware F$_1$ rewards. We go further on several fronts. Their rewards rely on ground-truth-guided greedy matching and require calling SAM to score points against masks, whereas our count reward uses a one-to-one Hungarian assignment, charges every duplicate as an explicit false positive, and requires no auxiliary model; this completely removes the need for NMS rather than merely mitigating it. Their outputs stop at points and boxes, whereas our chain-of-perception cascades from pointing to segmentation through detection. Finally, with $5\times$ fewer parameters, Falcon Perception-HD largely outperforms Rex-Omni on PBench, especially on very dense scenes (box F$_1$ $73.9$ vs $37.4$ on the dense split, Appendix~\ref{app:external_bench}). \citet{pinto2023tuning} argued more broadly
for aligning vision models with set-matching metrics rather than per-token
likelihood, an observation aligning with our results. Earlier
RL-for-localization work~\citep{caicedo2015active,mathe2016rlobject} cast
bounding-box prediction as a sequential decision process but predates 
autoregressive models and was confined to single-object regimes. The most conceptually aligned predecessor to our work is \citet{pinto2023tuning}, who demonstrated that autoregressive vision models (e.g., UViM~\cite{kolesnikov2022uvim}) can be aligned with non-differentiable set-matching metrics using REINFORCE \cite{williams1992simple}, directly addressing the limitations of per-token maximum likelihood. While their findings motivate our approach, their methodology relies on fully tokenized outputs and standard policy gradients, limiting scalability. Furthermore, none of these recent or foundational works address extreme dense perception (hundreds of instances per image), the complexities of a heterogeneous, multi-head action space, or the interaction between positive and existence calibration in the open-vocabulary setting. To resolve the mismatch between autoregressive generation and order-invariant evaluation, our reward design (\S\ref{sec:reward}) inherits the structural insight of DETR~\citep{carion2020detr,zhu2021deformabledetr,zhang2023dino}. By treating detection as a set-to-set matching problem, we use the Hungarian assignment algorithm to obtain a tractable, order-invariant alignment that explicitly trades off precision for recall, thereby closing the gap left by MLE.


\vspace{-1em}
\section{Method}
\label{sec:method}

\vspace{-1em}


We build our reinforcement learning framework upon Falcon Perception~\cite{bevli2026falcon}, an autoregressive multimodal model designed for open-vocabulary segmentation. For RL post-training, the critical architectural feature of this model is its structured, multi-head decoding interface. To detect and segment objects, the model emits a repeating sequence, referred to as the \emph{chain-of-perception}. For each detected instance, the model predicts the fixed sub-sequence: \begin{equation} \underbrace{w_1, \ldots, w_k}{\text{LM tokens}} ;\to; \underbrace{\texttt{},(x, y)}{\text{center}} ;\to; \underbrace{\texttt{},(h, w)}{\text{box size}} ;\to; \underbrace{\texttt{},m}{\text{mask}} ;\to; \cdots \label{eq:chain} \end{equation} Crucially, the tokens in this sequence are generated by different specialized heads operating over different vocabularies. Standard textual tokens such as the referring expression are sampled from a standard LM head. However, the center coordinates $(x,y)$ and bounding box dimensions $(h,w)$ are generated as categorical samples over discretized spatial bins by two independent, lightweight heads. Finally, the segmentation mask is generated deterministically by a parallel continuous head that computes the sigmoid of the inner product between the dense image features and the emitted \texttt{} token.

The action space of the policy $\pi_\theta$ is therefore heterogeneous. Depending on the sequence position in Eq.~\ref{eq:chain}, an action step requires a categorical sample over text tokens, coordinate bins, or size bins. This structural separation dictates how we must aggregate log-probabilities across distinct heads and decide which heads to explore during sampling.

\subsection{RL objective for multi-head autoregressive perception}
\label{sec:rl_objective}

The objective is to maximize the expected reward
\begin{equation}
\max_\theta \; \mathbb{E}_{o \sim \pi_\theta(\cdot \mid I, q)} \bigl[ r\bigl(\text{Set}(\hat{\mathcal{B}}(o)),\, \text{Set}(\mathcal{B}^*)\bigr) \bigr],
\label{eq:rl_objective}
\end{equation}
where each rollout $o = a_{1:T}$ is a complete response sampled from $\pi_\theta$, $\hat{\mathcal{B}}(o) = \{(x_i, y_i, h_i, w_i, m_i)\}_i$ is the set of object instances decoded from $o$ via the chain-of-perception of Eq.~\ref{eq:chain}, $\mathcal{B}^*$ is the corresponding ground-truth set, and $r(\cdot)$ is a single scalar set-matching reward (\S\ref{sec:reward}).

\paragraph{Group-relative advantages.} For each prompt $(I, q)$ we sample $G = 8$ rollouts $\{o_i\}_{i=1}^G \sim \pi_{\theta}$, compute their rewards $\{r_i\}$, and standardize within the group:
\begin{equation}
\hat{A}_i \;=\; \frac{r_i - \mu_G}{\sigma_G + \epsilon}, \qquad \mu_G = \tfrac{1}{G}\sum_j r_j, \quad \sigma_G^2 = \tfrac{1}{G}\sum_j (r_j - \mu_G)^2.
\label{eq:advantage}
\end{equation}
The advantage $\hat{A}_i$ measures the relative reward of rollout $i$ against its siblings on the same prompt. The group means serve as a learned baseline, reducing variance without a separate value function.

\vspace{-1em}

\paragraph{Multi-head log-probability.} The policy gradient is weighted by the score function $\nabla_\theta \log \pi_\theta(a_t \mid s_t)$, which decomposes across heads because the action space is heterogeneous:
\begin{equation}
\begin{aligned}
\log \pi_\theta(a_t \mid s_t) =& \log \pi_\text{LM}(w_t \mid h_t) \\
&+\; \mathbf{1}[w_t = \texttt{<coord>}]\,\log \pi_\text{coord}(x_t, y_t \mid h_t) \\
&+\; \mathbf{1}[w_t = \texttt{<size>}]\,\log \pi_\text{size}(h^b_t, w^b_t \mid h_t).
\end{aligned}
\label{eq:multi_head_logprob}
\end{equation}

The segmentation head is deterministic and does not contribute a REINFORCE term; it is therefore frozen during RL post-training. Of the three remaining heads (LM, coord, size), we sample only the LM and coordinate heads at training temperature, and decode the size head greedily. The size head was trained by SFT to predict box dimensions conditional on a correct center via teacher-forcing; once RL improves the coordinate head, the size head receives better conditioning at inference time and produces better boxes without a direct reward signal. We refer to this as the \emph{cascade effect} and validate it in \S\ref{sec:cascade_results}, where sampling the size head also yields no additional gain.

\vspace{-1em}

\paragraph{On-policy updates with engine-mismatch importance sampling.} We perform a single gradient update per rollout group, so the rollout and training policies share the same parameter checkpoint. Rollouts are nevertheless not generated by the same code path as the training-time forward pass: sampling uses a paged-attention inference engine optimized for long autoregressive decoding, while gradients are computed by torchtitan~\cite{liang2024torchtitan} training engine forward pass over packed sequences. The two engines use different attention kernels and accumulate floating-point error differently, which produces a small but nonzero discrepancy between the rollout-time log-probability $\log \pi_{\theta_\text{rollout}}$ and the training-time recomputation $\log \pi_\theta$ evaluated at the same parameters. Following the VeRL~\cite{sheng2024hybridflow} implementation, we absorb this discrepancy with a \emph{detached} importance sampling ratio
\begin{equation}
w^{(i)}_t \;=\; \mathrm{sg}\!\left(\frac{\pi_\theta(a^{(i)}_t \mid s^{(i)}_t)}{\pi_{\theta_\text{rollout}}(a^{(i)}_t \mid s^{(i)}_t)}\right),
\label{eq:is_weight}
\end{equation}
where $\mathrm{sg}(\cdot)$ denotes stop-gradient. The per-token loss is then
\begin{equation}
\ell^{(i)}_t \;=\; -\, w^{(i)}_t \cdot \hat{A}_i \cdot \log \pi_\theta(a^{(i)}_t \mid s^{(i)}_t),
\label{eq:loss_is}
\end{equation}
which corrects the expectation in Eq.~\ref{eq:rl_objective} for the engine mismatch without contributing additional gradient through the ratio itself. We tested PPO-style clipping of $w^{(i)}_t$  and observed no improvement.

\vspace{-1em}

\paragraph{Length-unbiased loss aggregation (Dr.\ GRPO).} We retain GRPO's~\cite{shao2024deepseekmath} group-standardized advantages (Eq.~\ref{eq:advantage}) and only modify how per-token losses are aggregated into a scalar. Standard GRPO normalizes each rollout's contribution by its own response length, $\tfrac{1}{G}\sum_i \tfrac{1}{T_i}\sum_t \ell^{(i)}_t$. Under this aggregation, a token in a short rollout receives $T_\text{long}/T_\text{short}$ times as much gradient as a token in a long rollout at equal advantage, which biases the policy towards shorter responses. In our setting, a single positive rollout can exceed $3 \times 10^4$ tokens, whereas a less complete one is an order of magnitude shorter, so the bias systematically discourages predicting more objects. We therefore replace the per-rollout $1/T_i$ factor with the Dr.\ GRPO~\cite{liu2025drgrpo} aggregation, which divides by a fixed constant $B \cdot T_\text{max}$ independent of the realized rollout lengths:
\begin{equation}
\mathcal{L}_\text{PG}(\theta) \;=\; \mathbb{E}_{q,I}\!\left[\frac{1}{B \cdot T_\text{max}} \sum_{i=1}^G \sum_{t=1}^{T_i} \ell^{(i)}_t\right].
\label{eq:dr_grpo}
\end{equation}
Each token's gradient contribution is now independent of the length of the rollout it belongs to.

\vspace{-1em}

\paragraph{Regularization.} We tested KL regularization against the SFT reference policy and per-head entropy bonuses, neither of which produced consistent reward improvements in our setting. To preserve exploration without explicit entropy bonuses, we instead use Clip-Cov~\cite{cui2025entropy}, which zeroes the policy-gradient contribution of the small fraction of tokens whose log-probability and advantage are most strongly correlated. These tokens are the ones most aggressively reinforced by the standard objective and the main drivers of premature distribution sharpening; suppressing them at a small selection ratio (we use $0.0002$) keeps entropy from collapsing while leaving the rest of the gradient untouched. We apply Clip-Cov per head with separate covariance bounds, $[1, 5]$ for the LM head and $[10, 50]$ for the coordinate head, chosen by inspecting the range of the maximum log-probability/advantage covariance for each head over early training.

\subsection{Reward Function}
\label{sec:reward}

\vspace{-1em}

The reward measures how well the predicted set of objects matches the ground-truth set. Let $\hat{\mathcal{B}}$ and $\mathcal{B}^\star$ denote the predicted and ground-truth instances, each represented in normalized $(x, y, h, w)$ center format with $n_\text{pred}$ and $n_\text{gt}$ elements, respectively. We pair predictions to ground truth by Hungarian assignment on the IoU matrix and accept only pairs whose IoU exceeds $\tau = 0.5$. The number of accepted pairs is $\mathrm{TP}$; unmatched predictions are false positives ($\mathrm{FP} = n_\text{pred} - \mathrm{TP}$), unmatched ground-truth instances are false negatives ($\mathrm{FN} = n_\text{gt} - \mathrm{TP}$). We define
\begin{equation}
r_\text{count} \;=\; -\bigl(\mathrm{FN} \;+\; \alpha \cdot \mathrm{FP}\bigr),
\label{eq:count_reward}
\end{equation}
with $\alpha = 0.3$. The reward counts how many ground-truth instances are missed and weights spurious predictions by $\alpha$, with no further dependence on the localization quality of accepted matches: above the IoU threshold, all matched pairs contribute equally. The asymmetry between $\mathrm{FN}$ (weight one) and $\mathrm{FP}$ (weight $\alpha$) reflects that recall is the harder failure mode in the dense regime, where the model tends to under-predict; we found stronger penalties on $\mathrm{FP}$ not to work as well.
We also experimented with finer-grained rewards that score localization quality within the matched set, e.g., the F1 score, which combines precision and recall, and the panoptic-quality reward of \cite{pinto2023tuning}, which weights each true positive by its IoU. Neither produced consistent improvements over Eq.~\ref{eq:count_reward}.

\subsection{Stop-gradient on existence tokens}
\label{sec:selective_grad}

\vspace{-0.7em}

After the last token of the input prompt, the LM head emits one of two markers that announce whether the referred concept is present in the scene: \texttt{<object\_found>} or \texttt{<no\_object\_found>}. The remainder of the response is then either a sequence of detected objects (positive case) or the end-of-sequence token (negative case). We exclude the log-probability of these two existence markers from the policy gradient by applying a stop-gradient at those positions.

The reason is that we run RL only on positive queries: negative queries carry no useful gradient signal beyond the binary existence decision, which RL cannot improve through sampling. On every positive rollout, the existence token is by construction \texttt{<object\_found>}, so keeping its log-probability in the policy gradient simply reinforces this token at every step, regardless of whether the rollout earned a high or low reward. The model gradually loses the ability inherited from SFT to discriminate scenes where the referred object is absent, and MCC collapses. Stopping the gradient at these two tokens prevents this drift while allowing the rest of the LM gradient to continue refining the features that feed into the existence head. We show in \S\ref{sec:existence_results} that this preserves, and in fact improves, MCC even though the policy never sees a negative query during RL. Further details in App.~\ref{app:E3}.

\section{Data Annotation Pipeline}
\label{sec:data_annotation}

\vspace{-1em}

We carefully design our data annotation pipeline to yield high-quality, informative training signals for both dense scenes and referring expression segmentation. Specifically, we use the base model's own predictions, which aligns with the idea of having the model reflect on its predictions during training.

\textbf{Dense scenes:} To evaluate and train on highly crowded environments, our objective is to curate scenes containing upwards of 200 instances. We initially generate pseudo-labels using Falcon Perception across 2.5k curated dense images. Although Falcon Perception surpasses the strict 200-instance limit of SAM3, the raw outputs still exhibit typical failure modes: duplicated masks, fragmented predictions for a single entity, aggregated masks spanning multiple distinct instances, and entirely overlooked regions. Some of these can be observed in the top rows of Fig.~\ref{fig:teaser} and supplementary Figs.~\ref{fig:app_dense_1} and~\ref{fig:app_dense_2}. Consequently, we employ human annotators to refine and bootstrap these initial predictions to guarantee high fidelity. 

\begin{figure}[t]
    \centering
    \includegraphics[page=1, width=\textwidth, trim={0 1.5cm 0 0}, clip]{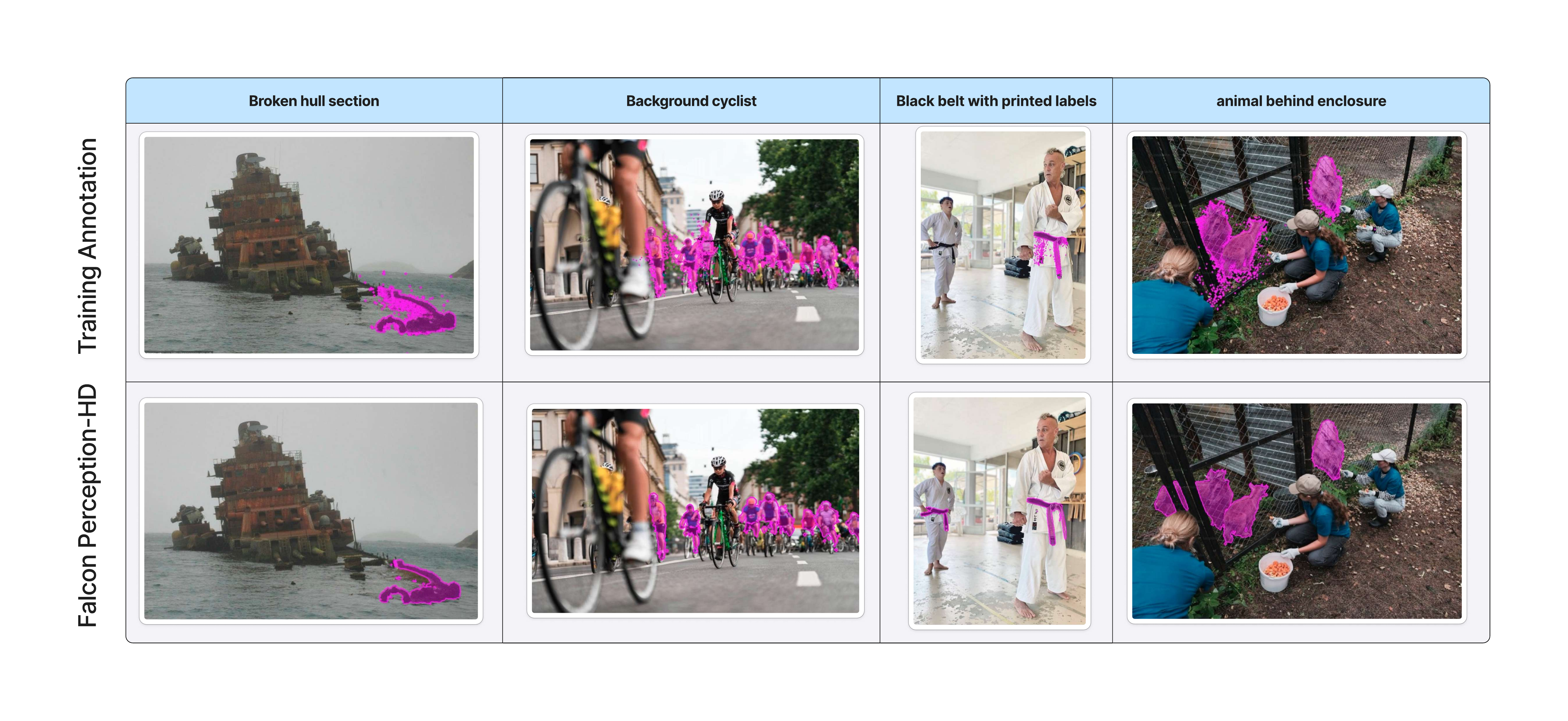}
    \caption{\textbf{Predictions exceed the annotations.} Original training annotations (top) vs.\ Falcon Perception-HD predictions (bottom).}
    \label{fig:annot_vs_fphd}
\end{figure}

\textbf{Hard referring expression samples:} To construct a dataset that provides a robust learning signal for our RL pipeline, we exploit the sampling capabilities of Falcon Perception. We perform inference on 200k image-expression pairs, generating one greedy prediction and a sampled set at pass@8 for each pair. By measuring the agreement across these rollouts, we observe two primary regimes: 1. When all predictions match, the model is highly confident; 2. when there is high variance among the rollouts, the model is uncertain. To provide a meaningful learning signal during RL post-training, we must retain only the high-variance predictions: RL relies on the ability to rank predictions within a group; indeed, if there is no reward variance, the advantages collapse to zero, and there is no gradient signal. Hence, we only focus on these high-variance samples. We provide some samples of these automatically annotated images in Fig.~\ref{fig:annot_vs_fphd}'s top row. 

In the high-variance regime, we utilize GPT-5 as a judge to evaluate the correctness of the generated candidates. We deliberately discard samples where the greedy prediction is correct. Instead, we isolate instances where the greedy prediction fails, but at least one sampled rollout succeeds. This filtering strategy yields a curated dataset with 11k samples, each with guaranteed high entropy, providing an optimal and challenging learning signal for RL training.

\vspace{-1em}
\section{Experiments}
\label{sec:experiments}

\definecolor{headergray}{RGB}{235,235,240}
\definecolor{ourrow}{RGB}{232,243,255}
\definecolor{deltarow}{RGB}{233,247,233}
\definecolor{collapsered}{RGB}{253,235,235}

\vspace{-1em}

We use Muon (\cite{jordan2024muon}) for 600 steps at a learning rate $10^{-5}$ with linear decay towards $10^{-6}$. Training is done on 64 A100 GPUs, with a global batch size of 64. All the training and implementation details are provided in the supplementary. 

We evaluate on two benchmarks: PBench~\cite{bevli2026falcon}, a referring expression segmentation benchmark with five difficulty levels (level-0 to level-4, ranging from short noun phrases to long compositional expressions) and a dedicated dense split where each image contains up to 500 instances of the queried category; and SACO~\cite{carion2025sam}, a seven-split benchmark (attributes, crowded, food, metaclip, sa1b, sport, wiki\_common). PBench is evaluated using COCO-style macro-F1, averaging over thresholds from $0.5$ to $0.95$, except on the dense split, where we use $0.5$. For SACO, which contains negative queries, we report macro-F1, pmF1, and MCC between $-1$ and $1$, measuring the model's ability to determine whether the concept exists in the scene.

\subsection{Main Results}
\label{sec:dense_results}

\vspace{-0.6em}

\begin{figure}[t]
    \centering
    \includegraphics[page=1, width=\textwidth]{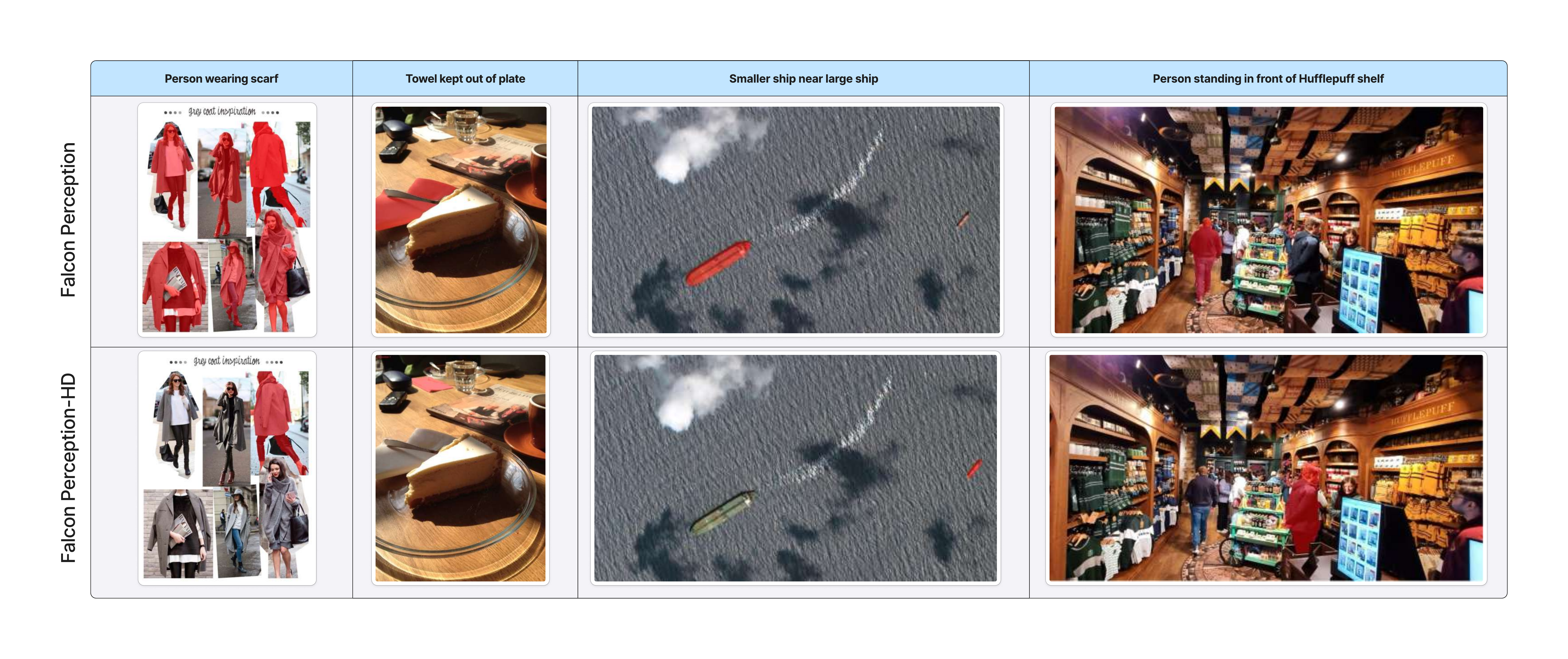}
    \caption{\textbf{PBench level~4 (relation binding) qualitative results.} Falcon Perception (top) vs.\ Falcon Perception-HD (bottom). RL improves on the hardest compositional referring expressions, where the SFT-only model tends to under/over-segment, miss the correct relational referent, or hallucinate.}
    \label{fig:level_4_qualitatives}
\end{figure}

Table~\ref{tab:pbench_refcoco} reports a state-of-the-art comparison on PBench. We observe improvements across all levels of difficulty; for instance, on level 4 (the most difficult relational expressions), Falcon Perception-HD improves by $2.7$ F$_1$ points, surpassing generalist VLMs like Qwen3-VL-8B that are more than $10\times$ larger. In Fig.~\ref{fig:level_4_qualitatives}, we show image-query pairs demonstrating how RL resolves these hard referring expressions (additional qualitative examples across levels 0–4 are provided in Figs.~\ref{fig:app_level4_more} and~\ref{fig:app_mix} of the appendix). Overall, the model improves by an average of $2.6$ points, establishing Falcon Perception-HD as the new state-of-the-art on this benchmark. Additional comparisons on external dense benchmarks (COCO-dense, LVIS-dense, and Dense200) are provided in Appendix~\ref{app:external_bench}.

Table~\ref{tab:saco_results_booktabs} presents results on SACO. Falcon Perception-HD improves upon the base model by $1.7$ points in macro-F1 and pmF1, and by $2\%$ in MCC. As described below, the main source of performance on this benchmark is MCC, because if we sample queries correctly as positives more often, macro-F1 and pmF1 will naturally increase. MCC is not a metric that can be improved via RL, as it is a classification problem; we still observe improvements in this metric after RL, which we believe is due to improved numerical stability from entropy reduction. Overall, Falcon Perception-HD sets a new state-of-the-art in macro-F1 and closes the gap with SAM3 in pmF1, while improving the base model's good object discrimination capabilities.  

\begin{table*}[t]
    \centering
    \small
    \setlength{\tabcolsep}{3pt} 

    \begin{minipage}[t]{0.68\textwidth}
        \caption{\textbf{PBench results (mask F$_1$).} COCO-style macro-F$_1$ averaged over IoU thresholds $0.5{:}0.95$ on levels 0--4, and F$_1$@0.5 on the dense split; all models are evaluated with their default post-processing. Avg is the unweighted mean over the six splits. $^*$ Reproduced results. Falcon Perception-HD outperforms all baselines on average, primarily driven by massive gains on the Dense split.}
        \label{tab:pbench_refcoco}
        \centering
        \renewcommand{\arraystretch}{1.15}
        \adjustbox{max width=\linewidth}{
        \begin{tabular}{lcccccccccc}
        \toprule
        \rowcolor{headergray}
        \textbf{Benchmark} & \rotatebox{65}{\textbf{Qwen3-2B}} & \rotatebox{65}{\textbf{Qwen3-4B}} & \rotatebox{65}{\textbf{Qwen3-8B}} & \rotatebox{65}{\textbf{Qwen3-30B}} & \rotatebox{65}{\textbf{MD2-2B}} & \rotatebox{65}{\textbf{MD3-9B}} & \rotatebox{65}{\textbf{SAM 3-0.9B}} & \rotatebox{65}{\textbf{FP$^*$}} & \rotatebox{65}{\textbf{FP-HD (Ours)}} & \rotatebox{65}{\textbf{$\Delta$}}\\
        \midrule
        \multicolumn{11}{l}{\textbf{PBench}} \\
        \quad L0: Simple objects      & 54.1 & 67.3 & 65.6 & \textbf{69.2} & \textit{68.0} & 63.2 & 64.3 & 63.7 & 64.9 & \textcolor{ForestGreen}{+1.1} \\
        \quad L1: Attribute           & 50.1 & 63.9 & \textit{65.6} & \textbf{68.8} & 58.4 & 59.6 & 54.4 & 63.8 & 64.2 & \textcolor{ForestGreen}{+0.5} \\
        \quad L2: OCR guided          & 41.2 & \textit{58.2} & \textit{58.2} & \textbf{61.2} & 46.6 & 41.8 & 24.6 & 38.3 & 40.4 & \textcolor{ForestGreen}{+2.1} \\
        \quad L3: Spatial understand. & 35.1 & 49.0 & 49.1 & 52.9 & 43.8 & 45.4 & 31.6 & \textit{53.4} & \textbf{54.7} & \textcolor{ForestGreen}{+1.3} \\
        \quad L4: Relation binding    & 36.7 & 48.3 & 50.9 & \textbf{55.2} & 36.9 & 42.3 & 33.3 & 49.1 & \textit{51.8} & \textcolor{ForestGreen}{+2.7} \\
        \quad Dense                   & 4.6  & 8.4  & 4.7  & 8.9  & 14.2 & 12.9 & 58.4 & \textit{72.3} & \textbf{80.5} & \textcolor{ForestGreen}{+8.2} \\
        \midrule
        \rowcolor{ourrow}
        \quad Average                 & 37.0 & 49.2 & 49.0 & 52.7 & 44.7 & 50.5 & 44.4 & \textit{56.8} & \textbf{59.4} & \textcolor{ForestGreen}{\textbf{+2.6}} \\
        \bottomrule
        \end{tabular}
        }
    \end{minipage}
    \hfill
    \begin{minipage}[t]{0.29\textwidth}
        \caption{\textbf{Cascade-effect ablation.} Avg F$_1$ on PBench in a reduced-compute setting. Sampling only LM+coord matches LM+coord+size, proving improvements naturally cascade to the size head.}
        \label{tab:cascade_ablation}
        \centering
        \renewcommand{\arraystretch}{1.15}
        \adjustbox{max width=\linewidth}{
        \begin{tabular}{lc}
            \toprule
            \rowcolor{headergray}
            \textbf{Sampled heads} & \textbf{Avg F$_1$} \\
            \midrule
            Falcon Perc. (pre-RL)            & 56.8 \\
            \quad + RL, center only           & 57.8 \\
            \quad + RL, LM+center+size        & 58.1 \\
            \rowcolor{ourrow}
            \quad + RL, LM+center \textbf{(Ours)} & \textbf{58.2} \\
            \bottomrule
        \end{tabular}
        }
    \end{minipage}
\end{table*}

\begin{table}[t]
    \centering
    \small
    \caption{\textbf{SA-Co Benchmark Results.} Comparison of Falcon Perception against state-of-the-art open-vocabulary segmentation models. We report Macro F$_1$, Positive Micro F$_1$ (pmF$_1$), and Image-Level MCC across all splits. $^*$ reproduced results.}
    \label{tab:saco_results_booktabs}
    \setlength{\tabcolsep}{4pt}
    \renewcommand{\arraystretch}{1.15}
    \adjustbox{max width=\textwidth}{
    \begin{tabular}{l*{24}{c}}
    \toprule
    \rowcolor{headergray}
    & \multicolumn{3}{c}{\textbf{Average}} & \multicolumn{3}{c}{\textbf{Metaclip}} & \multicolumn{3}{c}{\textbf{SA-1B}} & \multicolumn{3}{c}{\textbf{Crowded}} & \multicolumn{3}{c}{\textbf{Food\&Drink}} & \multicolumn{3}{c}{\textbf{Sports Equip.}} & \multicolumn{3}{c}{\textbf{Attributes}} & \multicolumn{3}{c}{\textbf{Wiki-Common}} \\
    \rowcolor{headergray}
    {\textbf{Model}}  & F$_1$ & pmF$_1$ & MCC & F$_1$ & pmF$_1$ & MCC & F$_1$ & pmF$_1$ & MCC & F$_1$ & pmF$_1$ & MCC & F$_1$ & pmF$_1$ & MCC & F$_1$ & pmF$_1$ & MCC & F$_1$ & pmF$_1$ & MCC & F$_1$ & pmF$_1$ & MCC \\
    \midrule
    SAM 3~\cite{carion2025sam} & 62.3 & \textbf{66.1} & \textbf{0.81} & 56.0 & \textbf{58.6} & \textbf{0.81} & \textbf{68.3} & \textbf{62.6} & \textbf{0.86} & \textbf{62.7} & \textbf{67.7} & \textbf{0.90} & 58.1 & 67.3 & \textit{0.79} & 71.2 & \textbf{73.8} & \textbf{0.89} & 71.1 & \textit{72.0} & \textit{0.76} & 49.0 & 60.9 & \textit{0.66} \\
    Falcon Perception$^*$
    & \textit{67.0} & 62.2 & 0.60
    & 57.4 & 51.3 & 0.59
    & 62.5 & 50.2 & 0.72
    & 59.3 & 58.8 & 0.64
    & \textit{70.3} & \textit{68.4} & 0.58
    & \textbf{75.1} & \textit{73.0} & 0.71
    & \textit{79.3} & 70.9 & 0.58
    & 64.8 & \textit{63.0} & 0.35 \\
    \rowcolor{ourrow}
    \textbf{Falcon Perception-HD (Ours)}
    & \textbf{68.7} & \textit{63.9} & 0.62
    & \textbf{64.8} & \textit{56.3} & 0.66
    & \textit{63.4} & \textit{52.4} & 0.72
    & \textit{59.7} & \textit{59.3} & 0.67
    & \textbf{72.8} & \textbf{71.8} & 0.59
    & \textit{75.0} & 72.9 & 0.74
    & \textbf{79.9} & 71.2 & 0.60
    & 65.0 & \textbf{63.2} & 0.37 \\
    \bottomrule
    \end{tabular}
    }
\end{table}

\vspace{-1em}

\subsection{Unlocking High-Density Perception}
\vspace{-0.6em}

The most significant effect of RL post-training occurs in very dense scenes, where most perception models either degrade sharply or collapse. While the base Falcon Perception is already the best model on the PBench dense split, it still misses objects and occasionally collapses. Falcon Perception-HD essentially eliminates these failure modes, improving macro-F1 by $8.2$ points and achieving a scale of dense perception unmatched by prior systems (see Fig.~\ref{fig:teaser} and Figs.~\ref{fig:app_dense_1} and~\ref{fig:app_dense_2} of the appendix). As shown in Table~\ref{tab:density}, the RL gain grows monotonically with scene density, scaling from $+1.5$ (0–2 objects) to $+21.5$ ($300+$ objects). Note that Table~\ref{tab:density} and Table~\ref{tab:pbench_refcoco} report the same predictions under two aggregation schemes, by ground-truth object count and by PBench split respectively, which is why the two averages differ. This is driven by the policy gradient naturally suppressing rollouts that under-cover the scene (via negative within-group advantages), directly penalizing premature sequence termination without explicit heuristics (see Fig.~\ref{fig:app_rollouts}).

\begin{table}[t]
    \centering
    \small
    \caption{\textbf{PBench segmentation F$_1$ by ground-truth object count.} The improvement from RL post-training grows monotonically with scene density.}
    \label{tab:density}
    \setlength{\tabcolsep}{6pt}
    \renewcommand{\arraystretch}{1.15}
    \adjustbox{max width=\textwidth}{
    \begin{tabular}{lccccccccc|c}
    \toprule
    \rowcolor{headergray}
    \textbf{Model} & \textbf{0--2} & \textbf{2--5} & \textbf{5--10} & \textbf{10--20} & \textbf{20--50} & \textbf{50--100} & \textbf{100--200} & \textbf{200--300} & \textbf{300+} & \textbf{Avg} \\
    \midrule
    Falcon Perception           & 54.9 & 54.6 & 60.2 & 61.4 & 63.5 & 78.7 & 76.5 & 67.4 & 55.9 & 63.7 \\
    \rowcolor{ourrow}
    \textbf{Falcon Perception-HD}  & \textbf{56.4} & \textbf{55.8} & \textbf{60.9} & \textbf{63.8} & \textbf{65.6} & \textbf{81.2} & \textbf{81.7} & \textbf{79.5} & \textbf{77.3} & \textbf{69.1} \\
    \midrule
    \rowcolor{deltarow}
    $\Delta$ &
    \textcolor{ForestGreen}{\textbf{+1.5}} &
    \textcolor{ForestGreen}{\textbf{+1.2}} &
    \textcolor{ForestGreen}{\textbf{+0.6}} &
    \textcolor{ForestGreen}{\textbf{+2.4}} &
    \textcolor{ForestGreen}{\textbf{+2.1}} &
    \textcolor{ForestGreen}{\textbf{+2.5}} &
    \textcolor{ForestGreen}{\textbf{+5.2}} &
    \textcolor{ForestGreen}{\textbf{+12.2}} &
    \textcolor{ForestGreen}{\textbf{+21.5}} &
    \textcolor{ForestGreen}{\textbf{+5.5}} \\
    \bottomrule
    \end{tabular}
    }
\end{table}

\subsection{Learned NMS: RL removes the need for inference-time deduplication}
\label{sec:nms_results}

\vspace{-1em}

Falcon-Perception relies on two post-processing steps to produce clean detection sets: a center-deduplication pass that drops predicted objects whose coordinates fall within a tunable bin distance of an earlier prediction, and non-maximum suppression (NMS) that removes duplicate masks above an IoU threshold (usually 0.5). Both are heuristics with thresholds that have to be tuned, and both can destroy genuine detections, e.g., \ NMS suppresses true overlapping instances such as objects close in the image plane or aligned in depth. We show that RL post-training removes the need for both by producing a policy whose output distribution is already deduplicated. The mechanism is visible at the rollout level: some rollouts emit near-duplicate predictions at slightly perturbed coordinates, each duplicate is matched as a Hungarian false positive, and the policy gradient suppresses these rollouts via a negative within-group advantage. We illustrate this in Fig.~\ref{fig:app_rollouts} of the appendix. We measure the resulting behavior using the mask redundancy rate (MRR), defined as the fraction of predicted masks suppressed by greedy NMS at an IoU threshold of $\tau = 0.5$.

We provide two ablations: Table~\ref{tab:coord_dedup_pbench} strips coordinate deduplication on PBench: Falcon Perception loses $7.5$ F$_1$ on the dense split (and $1.4$ on average) because the base model genuinely relies on the coord-dedup heuristic to clean up its outputs, while Falcon Perception-HD is essentially constant ($-0.2$ on dense); Table~\ref{tab:saco_postproc} is the corresponding test on SACO, with both post-processing steps toggled. The MRR is the clearest evidence: with NMS active, the base model already has $5.7\%$ redundant masks, while Falcon Perception-HD has $0.7\%$; removing coord dedup pushes the base model to $39.2\%$, while Falcon Perception-HD only drifts to $2.1\%$. The collapse propagates to pmF1: removing both post-processing steps drops Falcon Perception by $12.0$ points (from $62.6$ to $50.7$), while Falcon Perception-HD loses only $0.4$ points (from $63.9$ to $63.5$). In Appendix~\ref{app:ar_vs_detr}, we contrast this with DETR-style models, which solve repetition natively via bipartite matching but, as we show with a pass@$k$ study on SAM~3, do not benefit from RL post-training.

\begin{table*}[t]
    \centering
    \small
    \setlength{\tabcolsep}{4pt}
    
    \begin{minipage}[t]{0.54\textwidth}
        \caption{\textbf{F$_1$ scores with/without coordinate deduplication.} Stripping coord dedup costs the SFT baseline $7.5$ F$_1$ points on the dense split, as it predicts redundant, overlapping centers for the same object. However, Falcon Perception-HD is essentially unchanged ($-0.2$); RL has naturally internalized the spatial suppression that the base model offloaded to post-processing heuristics.}
        \label{tab:coord_dedup_pbench}
        \centering
        \renewcommand{\arraystretch}{1.15}
        \adjustbox{max width=\linewidth}{
        \begin{tabular}{lccccccc|c}
        \toprule
        \rowcolor{headergray}
        \textbf{Model} & \textbf{Dedup} & \textbf{L0} & \textbf{L1} & \textbf{L2} & \textbf{L3} & \textbf{L4} & \textbf{Dense} & \textbf{Avg} \\
        \midrule
        Falcon Perc.           & \checkmark    & 63.7 & 63.8 & 38.3 & 53.4 & 49.1 & 72.3 & 56.8 \\
        Falcon Perc.           & $\times$      & 63.6 & 63.5 & 38.3 & 53.4 & 49.0 & 64.8 & 55.4 \\
        \rowcolor{collapsered}
        \multicolumn{2}{l}{$\Delta$ Falcon Perc.}
            & \textcolor{red}{$-0.2$} & \textcolor{red}{$-0.3$} & $0.0$ & $0.0$ & \textcolor{red}{$-0.1$}
            & \textcolor{red}{\textbf{$-7.5$}} & \textcolor{red}{\textbf{$-1.4$}} \\
        \midrule
        \textbf{FP-HD (Ours)}   & \checkmark    & \textbf{64.9} & \textbf{64.2} & \textbf{40.4} & \textbf{54.7} & \textbf{51.8} & \textbf{80.5} & \textbf{59.4} \\
        \textbf{FP-HD (Ours)}   & $\times$      & \textbf{64.8} & \textbf{64.3} & \textbf{40.4} & \textbf{54.7} & \textbf{51.8} & \textbf{80.2} & \textbf{59.4} \\
        \rowcolor{deltarow}
        \multicolumn{2}{l}{$\Delta$ FP-HD}
            & \textcolor{ForestGreen}{$-0.1$} & \textcolor{ForestGreen}{$+0.1$} & $0.0$ & $0.0$ & $0.0$
            & \textcolor{ForestGreen}{\textbf{$-0.2$}} & \textcolor{ForestGreen}{\textbf{$0.0$}} \\
        \bottomrule
        \end{tabular}
        }
    \end{minipage}
    \hfill
    \begin{minipage}[t]{0.44\textwidth}
        \caption{\textbf{SACO post-processing ablation.} pmF1 and mask redundancy rate (MRR, \%). RL post-training removes the dependence on both post-processing steps: stripping them drops Falcon Perception by $12$ pmF1 points (yielding $36.2\%$ redundancy), but barely affects Falcon Perception-HD ($-0.4$ pmF1, $2.1\%$ MRR).}
        \label{tab:saco_postproc}
        \centering
        \renewcommand{\arraystretch}{1.15}
        \adjustbox{max width=\linewidth}{
        \begin{tabular}{lcccc}
        \toprule
        \rowcolor{headergray}
        \textbf{Setting} & \multicolumn{2}{c}{\textbf{Falcon Perc.}} & \multicolumn{2}{c}{\textbf{FP-HD (Ours)}} \\
        \cmidrule(lr){2-3} \cmidrule(lr){4-5}
        & pmF1 $\uparrow$ & MRR $\downarrow$ & pmF1 $\uparrow$ & MRR $\downarrow$ \\
        \midrule
        Dedup + NMS & 62.2 & 5.7\% & \textbf{63.9} & \textbf{0.7\%} \\
        No NMS                              & 61.2 & ---   & \textbf{63.7} & ---   \\
        No dedup                      & 62.6  & \textcolor{red}{36.2\%} & \textbf{63.9}          & \textbf{2.1\%} \\
        No post-proc                  & \textcolor{red}{50.7} & ---  & \textbf{63.5} & ---   \\
        \midrule
        \rowcolor{deltarow}
        $\Delta$ default $\to$ none & \textcolor{red}{\textbf{$-12.0$}} & --- & \textcolor{ForestGreen}{\textbf{$-0.4$}} & --- \\
        \bottomrule
        \end{tabular}
        }
    \end{minipage}
\end{table*}

\subsection{Disentangling RL from the curated data}
\label{sec:disentangle}

\vspace{-0.6em}

Our RL data is partly bootstrapped from Falcon Perception itself (\S\ref{sec:data_annotation}), so we disentangle the contribution of the curated data from that of the RL objective. We run one epoch of SFT on the exact same curated data and compare it against RL, on PBench (Table~\ref{tab:disentangle_pbench}) and SACO (Table~\ref{tab:disentangle_saco}). SFT on this heavily dense data improves the dense split but degrades levels 1 and 2, and, more importantly, worsens over-generation: masks per image go from 182 to 248, MRR without deduplication jumps from $35.1\%$ to $46.4\%$, and stripping NMS and deduplication now costs $14.8$ dense F$_1$ points against $7.0$ for the base model. Post-training this SFT model with GRPO (SFT~$\to$~RL) not only improves all levels but entirely fixes the repetition issues that SFT worsened. Duplicate detections also translate into latency: the SFT model takes $55.8$s per dense image, compared to $33.6$s for Falcon Perception-HD. Hence, the curated data alone does not explain our results: RL post-trained models are the only ones unaffected by mask repetition, whether initialized from the base model or from the curated-data SFT.

\begin{table*}[t]
    \centering
    \small
    \caption{\textbf{Disentangling RL from the curated data, PBench.} F$_1$ per split, MRR with and without coordinate deduplication (IoU $\tau{=}0.5$), dense F$_1$ without NMS or deduplication and its gap $\Delta$ to the default setting, average predicted masks per image and per-image latency on the dense split. SFT on the curated data amplifies over-generation and the dependence on post-processing; only RL removes both.}
    \label{tab:disentangle_pbench}
    \setlength{\tabcolsep}{4pt}
    \renewcommand{\arraystretch}{1.15}
    \adjustbox{max width=\textwidth}{
    \begin{tabular}{lll ccccccc cc cc cc}
    \toprule
    \rowcolor{headergray}
    \textbf{Run} & \textbf{Data} & \textbf{Objective} & \textbf{L0} & \textbf{L1} & \textbf{L2} & \textbf{L3} & \textbf{L4} & \textbf{Dense} & \textbf{Avg} & \textbf{MRR} & \textbf{\shortstack{MRR\\(no dedup)}} & \textbf{\shortstack{Dense F$_1$\\(no post-proc)}} & \textbf{$\Delta$} & \textbf{\shortstack{Masks\\/img}} & \textbf{Lat.} \\
    \midrule
    SFT init (pre-RL)      & original & MLE  & 63.8 & 63.8 & 38.3 & 53.4 & 49.1 & 71.9 & 56.7 & 1.58\% & 35.1\% & 64.9 & \textcolor{red}{$-7.0$}  & 182 & 39.6s \\
    SFT, 1 epoch           & curated  & MLE  & 63.9 & 62.7 & 37.6 & 54.2 & 48.9 & 77.3 & 57.4 & 2.52\% & \textcolor{red}{46.4\%} & 62.5 & \textcolor{red}{$-14.8$} & 248 & 55.8s \\
    SFT $\to$ RL           & curated  & GRPO & 64.2 & 63.5 & 39.1 & 54.6 & 50.4 & \textbf{80.9} & 58.8 & 0.61\% & 5.13\% & \textbf{80.4} & $-0.5$ & 165 & 38.2s \\
    \rowcolor{ourrow}
    \textbf{FP-HD (Ours)}  & curated  & GRPO & \textbf{64.9} & \textbf{64.2} & \textbf{40.4} & \textbf{54.7} & \textbf{51.8} & 80.5 & \textbf{59.4} & \textbf{0.22\%} & \textbf{3.19\%} & 78.8 & $-1.7$ & \textbf{156} & \textbf{33.6s} \\
    \bottomrule
    \end{tabular}
    }
\end{table*}

\begin{table}[t]
    \centering
    \small
    \caption{\textbf{Disentangling RL from the curated data, SACO.} SFT on the curated data raises redundancy without deduplication from $34.3\%$ to $42.6\%$, while both RL post-trained models reduce it.}
    \label{tab:disentangle_saco}
    \setlength{\tabcolsep}{5pt}
    \renewcommand{\arraystretch}{1.15}
    \adjustbox{max width=\textwidth}{
    \begin{tabular}{ll cc cc cc}
    \toprule
    \rowcolor{headergray}
    \textbf{Run} & \textbf{Objective} & \textbf{Avg F$_1$} & \textbf{MRR} & \textbf{\shortstack{MRR\\(no dedup)}} & \textbf{\shortstack{F$_1$\\(no post-proc)}} & \textbf{$\Delta$} & \textbf{MCC} \\
    \midrule
    SFT init (pre-RL)      & MLE  & 67.0 & 5.73\% & 34.3\% & 65.5 & \textcolor{red}{$-1.5$} & 59.6 \\
    SFT, 1 epoch           & MLE  & 67.5 & 9.57\% & \textcolor{red}{42.6\%} & 67.1 & $-0.4$ & 58.2 \\
    SFT $\to$ RL           & GRPO & 68.3 & \textbf{0.74\%} & 11.4\% & 68.1 & $-0.2$ & 58.9 \\
    \rowcolor{ourrow}
    \textbf{FP-HD (Ours)}  & GRPO & \textbf{68.7} & 0.76\% & \textbf{1.7\%} & \textbf{68.6} & $\mathbf{-0.1}$ & \textbf{62.0} \\
    \bottomrule
    \end{tabular}
    }
\end{table}


\vspace{-1em}

\subsection{The cascade effect}
\label{sec:cascade_results}

\vspace{-0.6em}

Here, we verify our decision to sample rollouts and compute policy gradients only over text tokens and center coordinates, while greedily selecting bounding box sizes. In Table~\ref{tab:cascade_ablation} we compare three configurations: (i) \emph{center only}, where the LM head is decoded greedily and only the coordinate head is sampled; (ii) \emph{LM + center} (our default), where both the LM and the center heads are sampled, and the size head is greedy; and (iii) \emph{LM + center + size}, where all three discrete heads are sampled. To keep the comparison tractable, this ablation is run in a reduced-compute setting.

First, while sampling only the coordinate head provides a strong baseline by improving localization, jointly sampling the LM head yields further gains. This is because the LM head explicitly controls sequence termination; exploring its action space allows the policy to learn better stopping criteria and improve recall in dense scenes. Second, extending sampling to the size head yields no additional benefit: \emph{LM + coord + size} performs identically to \emph{LM + coord}. This confirms the cascading optimization effect discussed in \S\ref{sec:rl_objective}. Because the architecture conditions subsequent heads on prior spatial predictions, improved center localization tightens the hidden state read by the downstream heads. This dynamic is visible during training: the size-head entropy decays in tandem with the LM and coordinate heads despite receiving no direct policy-gradient updates (Fig.~\ref{fig:app_training_curves}).

Table~\ref{tab:entropy} quantifies this effect. RL reduces the coordinate-head entropy by $63\%$ and, although the size head is frozen and decoded greedily, its entropy drops by $21\%$ as well, even though it never receives gradients. The SFT row is a controlled comparison: same initialization, same curated data, only the objective differs. Entropy barely moves on the coordinate head ($-4\%$) and goes the wrong way on the size head ($+8\%$), so the convergence observed in Fig.~\ref{fig:app_training_curves} is attributable to the RL objective and not to the data. While entropy collapse is a failure mode to be mitigated in the reasoning and RLHF literature~\cite{cui2025entropy}, since it forecloses the exploration and output diversity desired for math or coding, we argue it is a desirable property in dense perception: the set-level reward under on-policy sampling concentrates the policy on a single correct layout, and this reduction in uncertainty is precisely the mechanism that resolves duplication and premature termination.

\begin{table}[t]
    \centering
    \small
    \caption{\textbf{Mean policy entropy of the coordinate and size heads.} The size head is frozen and decoded greedily, yet RL reduces its entropy through the cascade effect. One epoch of SFT on the same data leaves both entropies essentially unchanged.}
    \label{tab:entropy}
    \setlength{\tabcolsep}{8pt}
    \renewcommand{\arraystretch}{1.15}
    \begin{tabular}{lcc}
        \toprule
        \rowcolor{headergray}
        \textbf{Model} & \textbf{Coord head} & \textbf{Size head} \\
        \midrule
        SFT init (pre-RL)        & 1.431 & 2.197 \\
        SFT, 1 epoch, same data  & 1.370 ($-4\%$) & 2.365 ($+8\%$) \\
        \rowcolor{ourrow}
        \textbf{FP-HD (RL)}      & \textbf{0.526} ($\mathbf{-63\%}$) & \textbf{1.729} ($\mathbf{-21\%}$) \\
        \bottomrule
    \end{tabular}
\end{table}

Crucially, this cascade effect also improves the frozen continuous segmentation head. As shown in Fig.~\ref{fig:annot_vs_fphd}, Falcon Perception-HD generates higher-quality, contiguous masks than those present in its own rollout annotations, effectively filtering out the "salt and pepper" noise found in the ground truth. This demonstrates that reducing uncertainty at the start of the chain-of-perception (the object center) fundamentally stabilizes the entire spatial pipeline.

\subsection{Preserving existence calibration via gradient detachment}
\label{sec:existence_results}
\vspace{-0.6em}

In Table~\ref{tab:existence_ablation}, we present the ablation on detaching the existence-token logprobs from the policy gradient to preserve the SFT-learned existence boundary. We observe that without stop-gradient, RL collapses MCC across all SACO splits: the average drops from $59.6$ (pre-RL) to $38.8$. The collapse leads to a misleading rise in pmF1 ($62.2 \to 66.0$) and macro F$_1$ ($67.0 \to 73.3$) because the policy learns to emit ``object found'' indiscriminately, inflating positive-class scores and destroying the existence calibration that MCC actually measures. This is because the model becomes overoptimistic: it only sees positive samples during RL, and the existence token is consistently reinforced at the expense of the non-existence token. As a byproduct, pmF1 is artificially inflated because the model predicts masks on more samples. Detaching the existence-token logprobs removes this perturbation and even yields better MCC performance than the base model. Detachment reverses both effects simultaneously: MCC recovers and even improves above pre-RL on every split (average $59.6 \to 62.2$), while pmF1 and macro F$_1$ also rise but more modestly. This also highlights, as noted previously, that MCC is the main driver of performance on SACO: in this setting the base model is already good enough when it decides to predict masks. 

\begin{table}[t]
    \centering
    \small
    \caption{\textbf{Impact of existence-token detachment on MCC.} Without the stop-gradient on the ``object found''/``no object found'' tokens, RL collapses MCC across every split (the policy emits ``found'' indiscriminately, which inflates pmF1/macro-F$_1$ but destroys the existence calibration MCC measures). With detachment, MCC improves above the pre-RL baseline on every split.}
    \label{tab:existence_ablation}
    \setlength{\tabcolsep}{6pt}
    \renewcommand{\arraystretch}{1.15}
    \adjustbox{max width=\textwidth}{
    \begin{tabular}{lcccccccc|c}
    \toprule
    \rowcolor{headergray}
    \textbf{Model} & \textbf{Attributes} & \textbf{Crowded} & \textbf{Food} & \textbf{MetaCLIP} & \textbf{SA-1B} & \textbf{Sport} & \textbf{Wiki} & \textbf{Avg MCC} & \textbf{Avg pmF1} \\
    \midrule
    Falcon Perception (pre-RL)                & 57.8 & 63.9 & 58.3 & 59.0 & 71.5 & 71.5 & 35.2 & 59.6 & 62.2 \\
    \rowcolor{collapsered}
    \quad + RL, no detach                     & \textcolor{red}{38.5} & \textcolor{red}{37.3} & \textcolor{red}{31.8} & \textcolor{red}{39.6} & \textcolor{red}{56.8} & \textcolor{red}{49.2} & \textcolor{red}{18.6} & \textcolor{red}{\textbf{38.8}} & 66.0 \\
    \rowcolor{ourrow}
    \quad + RL, with detach \textbf{(ours)}   & \textbf{60.1} & \textbf{66.7} & \textbf{59.5} & \textbf{66.1} & \textbf{72.3} & \textbf{73.6} & \textbf{37.4} & \textbf{62.2} & 63.9 \\
    \bottomrule
    \end{tabular}
    }
\end{table}
\vspace{-1em}



\section{Conclusion}
\label{sec:conclusion}

\vspace{-1em}


In this paper, we presented Falcon Perception-HD, demonstrating that reinforcement learning can fundamentally align autoregressive perception models with order-invariant spatial metrics. Rather than treating RL simply as a minor post-processing step, we showed that it resolves structural flaws inherent in Maximum Likelihood training. By efficiently restricting action sampling to the discrete spatial heads and applying a set-matching reward, we showed that an autoregressive policy can internalize spatial suppression, thereby replacing fragile heuristics such as NMS and center deduplication. Most importantly, our framework unlocks robustness in extreme-density regimes (up to 500 instances), solving the sequence collapse typical for long-horizon spatial generation. Ultimately, Falcon Perception-HD suggests that post-training is not just for language reasoning; it is a critical, scalable pathway for building robust perception models that no longer rely on inference-time heuristics.

\bibliographystyle{plainnat}
\bibliography{references}

@article{guo2025deepseek,
  title={Deepseek-r1: Incentivizing reasoning capability in llms via reinforcement learning},
  author={Guo, Daya and Yang, Dejian and Zhang, Haowei and Song, Junxiao and Wang, Peiyi and Zhu, Qihao and Xu, Runxin and Zhang, Ruoyu and Ma, Shirong and Bi, Xiao and others},
  journal={arXiv preprint arXiv:2501.12948},
  year={2025}
}

@article{kolesnikov2022uvim,
  title={Uvim: A unified modeling approach for vision with learned guiding codes},
  author={Kolesnikov, Alexander and Susano Pinto, Andr{\'e} and Beyer, Lucas and Zhai, Xiaohua and Harmsen, Jeremiah and Houlsby, Neil},
  journal={Advances in Neural Information Processing Systems},
  volume={35},
  pages={26295--26308},
  year={2022}
}

@article{jordan2024muon,
  title={Muon: An optimizer for hidden layers in neural networks, 2024},
  author={Jordan, Keller and Jin, Yuchen and Boza, Vlado and Jiacheng, You and Cesista, Franz and Newhouse, Laker and Bernstein, Jeremy},
  journal={URL https://kellerjordan. github. io/posts/muon},
  volume={6},
  number={3},
  pages={4},
  year={2024}
}

@article{williams1992simple,
  title={Simple statistical gradient-following algorithms for connectionist reinforcement learning},
  author={Williams, Ronald J},
  journal={Machine learning},
  volume={8},
  number={3},
  pages={229--256},
  year={1992},
  publisher={Springer}
}

@article{cui2025entropy,
  title={The entropy mechanism of reinforcement learning for reasoning language models},
  author={Cui, Ganqu and Zhang, Yuchen and Chen, Jiacheng and Yuan, Lifan and Wang, Zhi and Zuo, Yuxin and Li, Haozhan and Fan, Yuchen and Chen, Huayu and Chen, Weize and others},
  journal={arXiv preprint arXiv:2505.22617},
  year={2025}
}

@article{pix2seq,
  title={Pix2Seq: A Language Modeling Framework for Object Detection},
  author={Chen, Ting and Saxena, Saurabh and Li, Lala and Fleet, David J and Hinton, Geoffrey},
  journal={International Conference on Learning Representations},
  year={2022}
}

@article{pbench,
  title={PBench: A Benchmark for Referring Expression Segmentation Across Difficulty Levels and Dense Scenes},
  author={},
  journal={arXiv preprint arXiv:2603.27365},
  year={2026},
  note={TODO: verify exact citation, authors, and venue}
}

@article{carion2025sam,
  title={Sam 3: Segment anything with concepts},
  author={Carion, Nicolas and Gustafson, Laura and Hu, Yuan-Ting and Debnath, Shoubhik and Hu, Ronghang and Suris, Didac and Ryali, Chaitanya and Alwala, Kalyan Vasudev and Khedr, Haitham and Huang, Andrew and others},
  journal={arXiv preprint arXiv:2511.16719},
  year={2025}
}

@inproceedings{lu2023unifiedio,
  title     = {Unified-{IO}: A Unified Model for Vision, Language, and
               Multi-modal Tasks},
  author    = {Lu, Jiasen and Clark, Christopher and Zellers, Rowan and
               Mottaghi, Roozbeh and Kembhavi, Aniruddha},
  booktitle = {International Conference on Learning Representations (ICLR)},
  year      = {2023}
}

@inproceedings{lu2024unifiedio2,
  title     = {Unified-{IO}~2: Scaling Autoregressive Multimodal Models with
               Vision, Language, Audio, and Action},
  author    = {Lu, Jiasen and Clark, Christopher and Lee, Sangho and Zhang,
               Zichen and Khosla, Savya and Marten, Ryan and Hoiem, Derek
               and Kembhavi, Aniruddha},
  booktitle = {Conference on Computer Vision and Pattern Recognition (CVPR)},
  year      = {2024}
}

@inproceedings{wang2022ofa,
  title     = {{OFA}: Unifying Architectures, Tasks, and Modalities Through a
               Simple Sequence-to-Sequence Learning Framework},
  author    = {Wang, Peng and Yang, An and Men, Rui and Lin, Junyang and Bai,
               Shuai and Li, Zhikang and Ma, Jianxin and Zhou, Chang and
               Zhou, Jingren and Yang, Hongxia},
  booktitle = {International Conference on Machine Learning (ICML)},
  year      = {2022}
}

@inproceedings{xiao2024florence2,
  title     = {{Florence-2}: Advancing a Unified Representation for a Variety
               of Vision Tasks},
  author    = {Xiao, Bin and Wu, Haiping and Xu, Weijian and Dai, Xiyang and
               Hu, Houdong and Lu, Yumao and Zeng, Michael and Liu, Ce and
               Yuan, Lu},
  booktitle = {Conference on Computer Vision and Pattern Recognition (CVPR)},
  year      = {2024}
}

@inproceedings{lai2024lisa,
  title     = {{LISA}: Reasoning Segmentation via Large Language Model},
  author    = {Lai, Xin and Tian, Zhuotao and Chen, Yukang and Li, Yanwei and
               Yuan, Yuhui and Liu, Shu and Jia, Jiaya},
  booktitle = {Conference on Computer Vision and Pattern Recognition (CVPR)},
  year      = {2024}
}

@inproceedings{rasheed2024glamm,
  title     = {{GLaMM}: Pixel Grounding Large Multimodal Model},
  author    = {Rasheed, Hanoona and Maaz, Muhammad and Shaji, Sahal and
               Shaker, Abdelrahman and Khan, Salman and Cholakkal, Hisham and
               Anwer, Rao M. and Xing, Eric and Yang, Ming-Hsuan and Khan,
               Fahad S.},
  booktitle = {Conference on Computer Vision and Pattern Recognition (CVPR)},
  year      = {2024}
}

@inproceedings{ren2024pixellm,
  title     = {{PixelLM}: Pixel Reasoning with Large Multimodal Model},
  author    = {Ren, Zhongwei and Huang, Zhicheng and Wei, Yunchao and Zhao,
               Yao and Fu, Dongmei and Feng, Jiashi and Jin, Xiaojie},
  booktitle = {Conference on Computer Vision and Pattern Recognition (CVPR)},
  year      = {2024}
}

@inproceedings{wu2024visionllmv2,
  title     = {{VisionLLM}~v2: An End-to-End Generalist Multimodal Large
               Language Model for Hundreds of Vision-Language Tasks},
  author    = {Wu, Jiannan and Zhong, Muyan and Xing, Sen and Lai, Zeqiang and
               Liu, Zhaoyang and Wang, Wenhai and Chen, Zhe and Zhu, Xizhou
               and Lu, Lewei and Lu, Tong and Luo, Ping and Li, Hongsheng and
               Dai, Jifeng},
  booktitle = {Advances in Neural Information Processing Systems (NeurIPS)},
  year      = {2024}
}

@article{bai2023qwenvl,
  title   = {{Qwen-VL}: A Versatile Vision-Language Model for Understanding,
             Localization, Text Reading, and Beyond},
  author  = {Bai, Jinze and Bai, Shuai and Yang, Shusheng and Wang, Shijie
             and Tan, Sinan and Wang, Peng and Lin, Junyang and Zhou,
             Chang and Zhou, Jingren},
  journal = {arXiv preprint arXiv:2308.12966},
  year    = {2023}
}

@article{wang2024qwen2vl,
  title   = {{Qwen2-VL}: Enhancing Vision-Language Model's Perception of the
             World at Any Resolution},
  author  = {Wang, Peng and Bai, Shuai and Tan, Sinan and Wang, Shijie and
             Fan, Zhihao and Bai, Jinze and Chen, Keqin and Liu, Xuejing
             and Wang, Jialin and Ge, Wenbin and Fan, Yang and Dang, Kai
             and Du, Mengfei and Ren, Xuancheng and Men, Rui and Liu,
             Dayiheng and Zhou, Chang and Zhou, Jingren and Lin, Junyang},
  journal = {arXiv preprint arXiv:2409.12191},
  year    = {2024}
}

@misc{qwen2025qwen3vl,
  title        = {{Qwen3-VL} Technical Report},
  author       = {{Qwen Team}},
  year         = {2025},
  howpublished = {\url{https://qwenlm.github.io/blog/qwen3-vl/}}
}

@misc{korrapati2024moondream,
  title        = {Moondream: A Tiny Vision Language Model},
  author       = {Korrapati, Vikhyat},
  year         = {2024},
  howpublished = {\url{https://moondream.ai}}
}

@article{bevli2026falcon,
  title   = {Falcon Perception},
  author  = {Bevli, Aviraj and Chaybouti, Sofian and Dahou, Yasser and Hacid,
             Hakim and Huynh, Ngoc Dung and Le Khac, Phuc H. and Narayan,
             Sanath and Para, Wamiq Reyaz and Singh, Ankit},
  journal = {arXiv preprint arXiv:2603.27365},
  year    = {2026},
  note    = {\url{https://huggingface.co/tiiuae/Falcon-Perception}}
}

@article{chaybouti2025siglino,
  title={SigLino: Efficient Multi-Teacher Distillation for Agglomerative Vision Foundation Models},
  author={Chaybouti, Sofian and Narayan, Sanath and Dahou, Yasser and Khac, Ph{\'u}c H L{\^e} and Singh, Ankit and Huynh, Ngoc Dung and Para, Wamiq Reyaz and Kuehne, Hilde and Hacid, Hakim},
  journal={arXiv preprint arXiv:2512.20157},
  year={2025}
}

@inproceedings{minderer2024owlv2,
  title     = {Scaling Open-Vocabulary Object Detection},
  author    = {Minderer, Matthias and Gritsenko, Alexey and Houlsby, Neil},
  booktitle = {Advances in Neural Information Processing Systems (NeurIPS)},
  year      = {2024}
}

@inproceedings{liu2024gdino,
  title     = {Grounding {DINO}: Marrying {DINO} with Grounded Pre-training
               for Open-Set Object Detection},
  author    = {Liu, Shilong and Zeng, Zhaoyang and Ren, Tianhe and Li, Feng
               and Zhang, Hao and Yang, Jie and Li, Chunyuan and Yang, Jianwei
               and Su, Hang and Zhu, Jun and Zhang, Lei},
  booktitle = {European Conference on Computer Vision (ECCV)},
  year      = {2024}
}

@article{ren2024dinox,
  title   = {{DINO-X}: A Unified Vision Model for Open-World Object Detection
             and Understanding},
  author  = {Ren, Tianhe and Chen, Yihao and Jiang, Qing and Zeng, Zhaoyang
             and Xiong, Yuda and Liu, Wenlong and Ma, Zhengyu and Shen, Junyi
             and Gao, Yuan and Jiang, Xiaoke and Chen, Xingyu and Song, Zhuheng
             and Zhang, Yuhong and Huang, Hongjie and Gao, Han and Liu, Shilong
             and Zhang, Hao and Li, Feng and Yu, Kent and Zhang, Lei},
  journal = {arXiv preprint arXiv:2411.14347},
  year    = {2024}
}

@inproceedings{shen2024ape,
  title     = {Aligning and Prompting Everything All at Once for Universal
               Visual Perception},
  author    = {Shen, Yunhang and Fu, Chaoyou and Chen, Peixian and Zhang,
               Mengdan and Li, Ke and Sun, Xing and Wu, Yunsheng and Lin, Shaohui
               and Ji, Rongrong},
  booktitle = {Conference on Computer Vision and Pattern Recognition (CVPR)},
  year      = {2024}
}

@inproceedings{kirillov2023sam,
  title     = {Segment Anything},
  author    = {Kirillov, Alexander and Mintun, Eric and Ravi, Nikhila and Mao,
               Hanzi and Rolland, Chloe and Gustafson, Laura and Xiao, Tete
               and Whitehead, Spencer and Berg, Alexander C. and Lo, Wan-Yen
               and Doll{\'a}r, Piotr and Girshick, Ross},
  booktitle = {International Conference on Computer Vision (ICCV)},
  year      = {2023}
}

@article{ravi2024sam2,
  title   = {{SAM}~2: Segment Anything in Images and Videos},
  author  = {Ravi, Nikhila and Gabeur, Valentin and Hu, Yuan-Ting and Hu,
             Ronghang and Ryali, Chaitanya and Ma, Tengyu and Khedr, Haitham
             and R{\"a}dle, Roman and Rolland, Chloe and Gustafson, Laura
             and Mintun, Eric and Pan, Junting and Alwala, Kalyan Vasudev and
             Carion, Nicolas and Wu, Chao-Yuan and Girshick, Ross and Doll{\'a}r,
             Piotr and Feichtenhofer, Christoph},
  journal = {arXiv preprint arXiv:2408.00714},
  year    = {2024}
}

@inproceedings{ouyang2022instructgpt,
  title     = {Training Language Models to Follow Instructions with Human
               Feedback},
  author    = {Ouyang, Long and Wu, Jeffrey and Jiang, Xu and Almeida, Diogo
               and Wainwright, Carroll L. and Mishkin, Pamela and Zhang, Chong
               and Agarwal, Sandhini and Slama, Katarina and Ray, Alex and
               others},
  booktitle = {Advances in Neural Information Processing Systems (NeurIPS)},
  year      = {2022}
}

@article{bai2022rlhf,
  title   = {Training a Helpful and Harmless Assistant with Reinforcement
             Learning from Human Feedback},
  author  = {Bai, Yuntao and Jones, Andy and Ndousse, Kamal and Askell, Amanda
             and Chen, Anna and DasSarma, Nova and Drain, Dawn and Fort,
             Stanislav and Ganguli, Deep and Henighan, Tom and others},
  journal = {arXiv preprint arXiv:2204.05862},
  year    = {2022}
}

@article{schulman2017ppo,
  title   = {Proximal Policy Optimization Algorithms},
  author  = {Schulman, John and Wolski, Filip and Dhariwal, Prafulla and
             Radford, Alec and Klimov, Oleg},
  journal = {arXiv preprint arXiv:1707.06347},
  year    = {2017}
}

@inproceedings{rafailov2023dpo,
  title     = {Direct Preference Optimization: Your Language Model is Secretly
               a Reward Model},
  author    = {Rafailov, Rafael and Sharma, Archit and Mitchell, Eric and
               Ermon, Stefano and Manning, Christopher D. and Finn, Chelsea},
  booktitle = {Advances in Neural Information Processing Systems (NeurIPS)},
  year      = {2023}
}

@article{lambert2024tulu3,
  title   = {{T{\"u}lu}~3: Pushing Frontiers in Open Language Model
             Post-Training},
  author  = {Lambert, Nathan and Morrison, Jacob and Pyatkin, Valentina and
             Huang, Shengyi and Ivison, Hamish and Brahman, Faeze and Miranda,
             Lester James V. and Liu, Alisa and Dziri, Nouha and Lyu, Shane
             and Gu, Yuling and Malik, Saumya and Graf, Victoria and Hwang,
             Jena D. and Yang, Jiangjiang and Bras, Ronan Le and Tafjord,
             Oyvind and Wilhelm, Chris and Soldaini, Luca and Smith, Noah A.
             and Wang, Yizhong and Dasigi, Pradeep and Hajishirzi, Hannaneh},
  journal = {arXiv preprint arXiv:2411.15124},
  year    = {2024}
}

@article{shao2024deepseekmath,
  title   = {{DeepSeekMath}: Pushing the Limits of Mathematical Reasoning in
             Open Language Models},
  author  = {Shao, Zhihong and Wang, Peiyi and Zhu, Qihao and Xu, Runxin and
             Song, Junxiao and Zhang, Mingchuan and Li, Y. K. and Wu, Y. and
             Guo, Daya},
  journal = {arXiv preprint arXiv:2402.03300},
  year    = {2024}
}

@article{yu2025dapo,
  title   = {{DAPO}: An Open-Source {LLM} Reinforcement Learning System at
             Scale},
  author  = {Yu, Qiying and Zhang, Zheng and Zhu, Ruofei and Yuan, Yufeng and
             Zuo, Xiaochen and Yue, Yu and Fan, Tiantian and Liu, Gaohong and
             Liu, Lingjun and Liu, Xin and Lin, Haibin and Lin, Zhiqi and Ma,
             Bole and Sheng, Guangming and Tong, Yuxuan and Zhang, Chi and
             Zhang, Mofan and Zhang, Wang and Zhu, Hang and Zhu, Jinhua and
             Chen, Jiaze and Chen, Jiangjie and Wang, Chengyi and Yu, Hongli
             and Dai, Weinan and Song, Yuxuan and Wei, Xiangpeng and Zhou,
             Hao and Liu, Jingjing and Ma, Wei-Ying and Zhang, Ya-Qin and
             Yan, Lin and Qiao, Mu and Wu, Yonghui and Wang, Mingxuan},
  journal = {arXiv preprint arXiv:2503.14476},
  year    = {2025}
}

@article{minimax2025cispo,
  title   = {{MiniMax-M1}: Scaling Test-Time Compute Efficiently with Lightning
             Attention},
  author  = {{MiniMax}},
  journal = {arXiv preprint arXiv:2506.13585},
  year    = {2025}
}

@article{liu2025drgrpo,
  title   = {Understanding {R1-Zero}-Like Training: A Critical Perspective},
  author  = {Liu, Zichen and Chen, Changyu and Li, Wenjun and Qi, Penghui and
             Pang, Tianyu and Du, Chao and Lee, Wee Sun and Lin, Min},
  journal = {arXiv preprint arXiv:2503.20783},
  year    = {2025}
}

@article{wang2025gtpo,
  title   = {{GTPO}: Trajectory-Based Policy Optimization in Large Language
             Models},
  author  = {Wang, Marco Simoni and others},
  journal = {arXiv preprint arXiv:2508.03772},
  year    = {2025}
}

@article{hu2025reinforcepp,
  title   = {{REINFORCE++}: A Simple and Efficient Approach for Aligning Large
             Language Models},
  author  = {Hu, Jian},
  journal = {arXiv preprint arXiv:2501.03262},
  year    = {2025}
}

@article{sheng2024hybridflow,
  title   = {HybridFlow: A Flexible and Efficient RLHF Framework},
  author  = {Guangming Sheng and Chi Zhang and Zilingfeng Ye and Xibin Wu and Wang Zhang and Ru Zhang and Yanghua Peng and Haibin Lin and Chuan Wu},
  year    = {2024},
  journal = {arXiv preprint arXiv: 2409.19256}
}

@article{liu2025visualrft,
  title   = {Visual-{RFT}: Visual Reinforcement Fine-Tuning},
  author  = {Liu, Ziyu and Sun, Zeyi and Zang, Yuhang and Dong, Xiaoyi and
             Cao, Yuhang and Duan, Haodong and Lin, Dahua and Wang, Jiaqi},
  journal = {arXiv preprint arXiv:2503.01785},
  year    = {2025}
}

@article{huang2025visionr1,
  title   = {{Vision-R1}: Incentivizing Reasoning Capability in Multimodal
             Large Language Models},
  author  = {Huang, Wenxuan and Jia, Bohan and Zhai, Zijie and Cao, Shaosheng
             and Ye, Zoey and Zhao, Fei and Hu, Yaoyao and Lin, Shaohui},
  journal = {arXiv preprint arXiv:2503.06749},
  year    = {2025}
}

@misc{chen2025r1v,
  title        = {{R1-V}: Reinforcing Super Generalization Ability in
                  Vision-Language Models with Less Than \$3},
  author       = {Chen, Liang and Li, Lei and Zhao, Haozhe and Song, Yifan
                  and Vinci},
  year         = {2025},
  howpublished = {\url{https://github.com/Deep-Agent/R1-V}}
}

@article{peng2025lmmr1,
  title   = {{LMM-R1}: Empowering 3B {LMMs} with Strong Reasoning Abilities
             Through Two-Stage Rule-Based {RL}},
  author  = {Peng, Yingzhe and Zhang, Gongrui and Zhang, Miaosen and You,
             Zhiyuan and Liu, Jie and Zhu, Qipeng and Yang, Kai and Xu, Xingzhong
             and Geng, Xin and Yang, Xu},
  journal = {arXiv preprint arXiv:2503.07536},
  year    = {2025}
}

@article{liu2025visionreasoner,
  title   = {{VisionReasoner}: Unified Visual Perception and Reasoning via
             Reinforcement Learning},
  author  = {Liu, Yuqi and Qu, Tianyuan and Zhong, Zhisheng and Peng, Bohao
             and Liu, Shu and Yu, Bei and Jia, Jiaya},
  journal = {arXiv preprint arXiv:2505.12081},
  year    = {2025}
}

@article{liu2025segzero,
  title   = {{Seg-Zero}: Reasoning-Chain Guided Segmentation via Cognitive
             Reinforcement},
  author  = {Liu, Yuqi and Peng, Bohao and Zhong, Zhisheng and Yue, Zihao and
             Lu, Fanbin and Yu, Bei and Jia, Jiaya},
  journal = {arXiv preprint arXiv:2503.06520},
  year    = {2025}
}

@inproceedings{pinto2023tuning,
  title={Tuning computer vision models with task rewards},
  author={Pinto, Andr{\'e} Susano and Kolesnikov, Alexander and Shi, Yuge and Beyer, Lucas and Zhai, Xiaohua},
  booktitle={International Conference on Machine Learning},
  pages={33229--33239},
  year={2023},
  organization={PMLR}
}

@inproceedings{caicedo2015active,
  title     = {Active Object Localization with Deep Reinforcement Learning},
  author    = {Caicedo, Juan C. and Lazebnik, Svetlana},
  booktitle = {International Conference on Computer Vision (ICCV)},
  year      = {2015}
}

@article{mathe2016rlobject,
  title   = {Reinforcement Learning for Visual Object Detection},
  author  = {Mathe, Stefan and Pirinen, Aleksis and Sminchisescu, Cristian},
  journal = {Conference on Computer Vision and Pattern Recognition (CVPR)},
  year    = {2016}
}

@inproceedings{carion2020detr,
  title     = {End-to-End Object Detection with Transformers},
  author    = {Carion, Nicolas and Massa, Francisco and Synnaeve, Gabriel and
               Usunier, Nicolas and Kirillov, Alexander and Zagoruyko, Sergey},
  booktitle = {European Conference on Computer Vision (ECCV)},
  year      = {2020}
}

@inproceedings{zhu2021deformabledetr,
  title     = {Deformable {DETR}: Deformable Transformers for End-to-End
               Object Detection},
  author    = {Zhu, Xizhou and Su, Weijie and Lu, Lewei and Li, Bin and Wang,
               Xiaogang and Dai, Jifeng},
  booktitle = {International Conference on Learning Representations (ICLR)},
  year      = {2021}
}

@inproceedings{zhang2023dino,
  title     = {{DINO}: {DETR} with Improved DeNoising Anchor Boxes for
               End-to-End Object Detection},
  author    = {Zhang, Hao and Li, Feng and Liu, Shilong and Zhang, Lei and
               Su, Hang and Zhu, Jun and Ni, Lionel M. and Shum, Heung-Yeung},
  booktitle = {International Conference on Learning Representations (ICLR)},
  year      = {2023}
}

@inproceedings{liang2024torchtitan,
  title={TorchTitan: One-stop PyTorch native solution for production ready LLM pre-training},
  author={Liang, Wanchao and Liu, Tianyu and Wright, Less and Constable, Will and Gu, Andrew and Huang, Chien-Chin and Zhang, Iris and Feng, Wei and Huang, Howard and Wang, Junjie and others},
  booktitle={International Conference on Learning Representations (ICLR)},
  year={2025}
}

@inproceedings{jiang2026detect,
  title={Detect anything via next point prediction},
  author={Jiang, Qing and Huo, Junan and Chen, Xingyu and Xiong, Yuda and Zeng, Zhaoyang and Chen, Yihao and Ren, Tianhe and Yu, Junzhi and Zhang, Lei},
  booktitle={Proceedings of the IEEE/CVF Conference on Computer Vision and Pattern Recognition},
  pages={25472--25483},
  year={2026}
}

@article{wang2026locateanything,
  title={LocateAnything: Fast and high-quality vision-language grounding with parallel box decoding},
  author={Wang, Shihao and Liu, Shilong and Kuang, Yuanguo and Wei, Xinyu and Liu, Yangzhou and Li, Zhiqi and Man, Yunze and Chen, Guo and Tao, Andrew and Liu, Guilin and others},
  journal={arXiv preprint arXiv:2605.27365},
  year={2026}
}

\newpage
\appendix

\section{Additional qualitative results}
\label{app:qualitatives}

\subsection{Dense scenes}
\label{app:dense_qualitatives}

\begin{figure}[h]
    \centering
    \includegraphics[page=1, width=\textwidth]{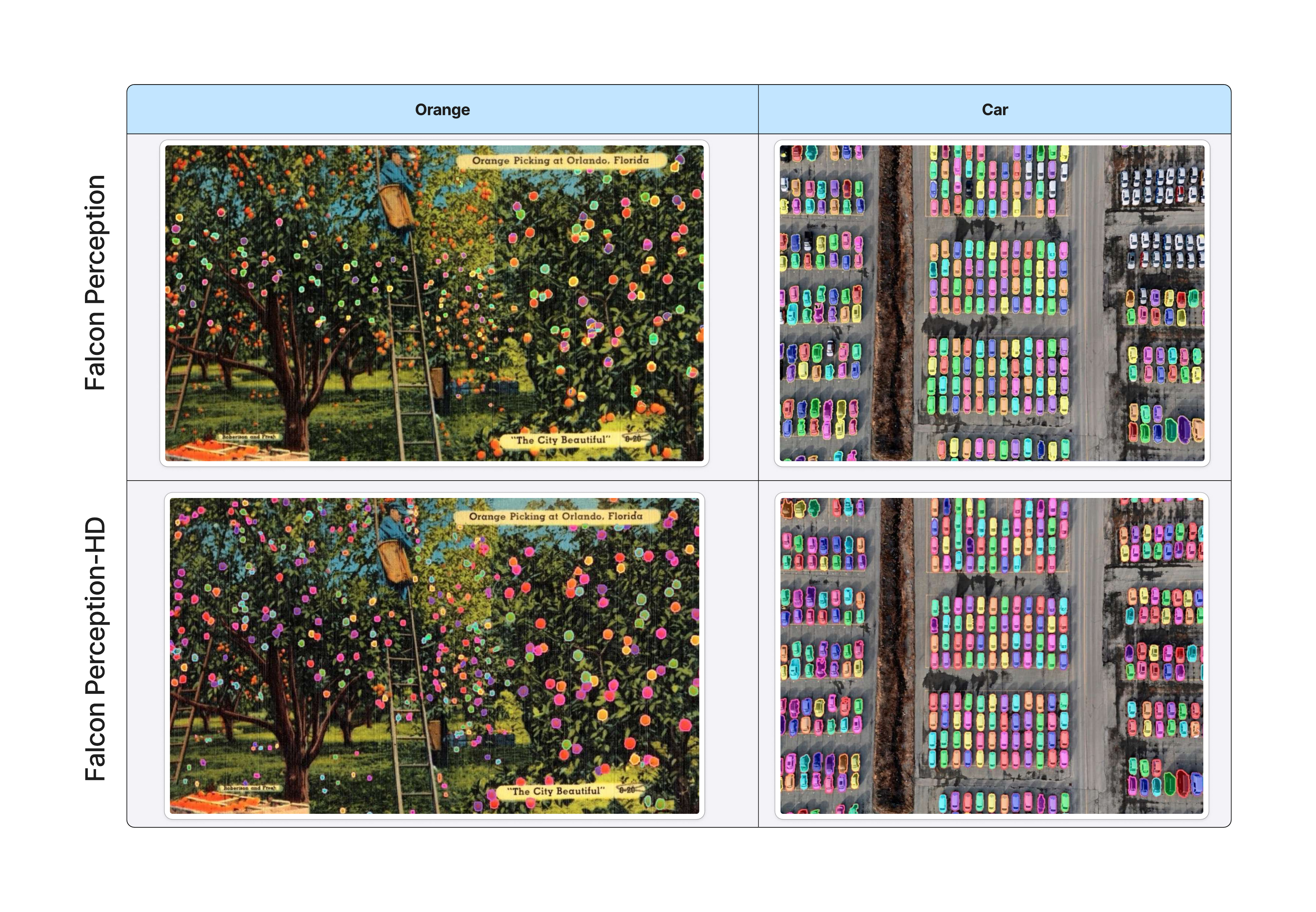}
    \caption{\textbf{Dense scenes, additional examples (1/2).} Top: Falcon Perception. Bottom: Falcon Perception-HD.}
    \label{fig:app_dense_1}
\end{figure}

\begin{figure}[h]
    \centering
    \includegraphics[page=1, width=\textwidth]{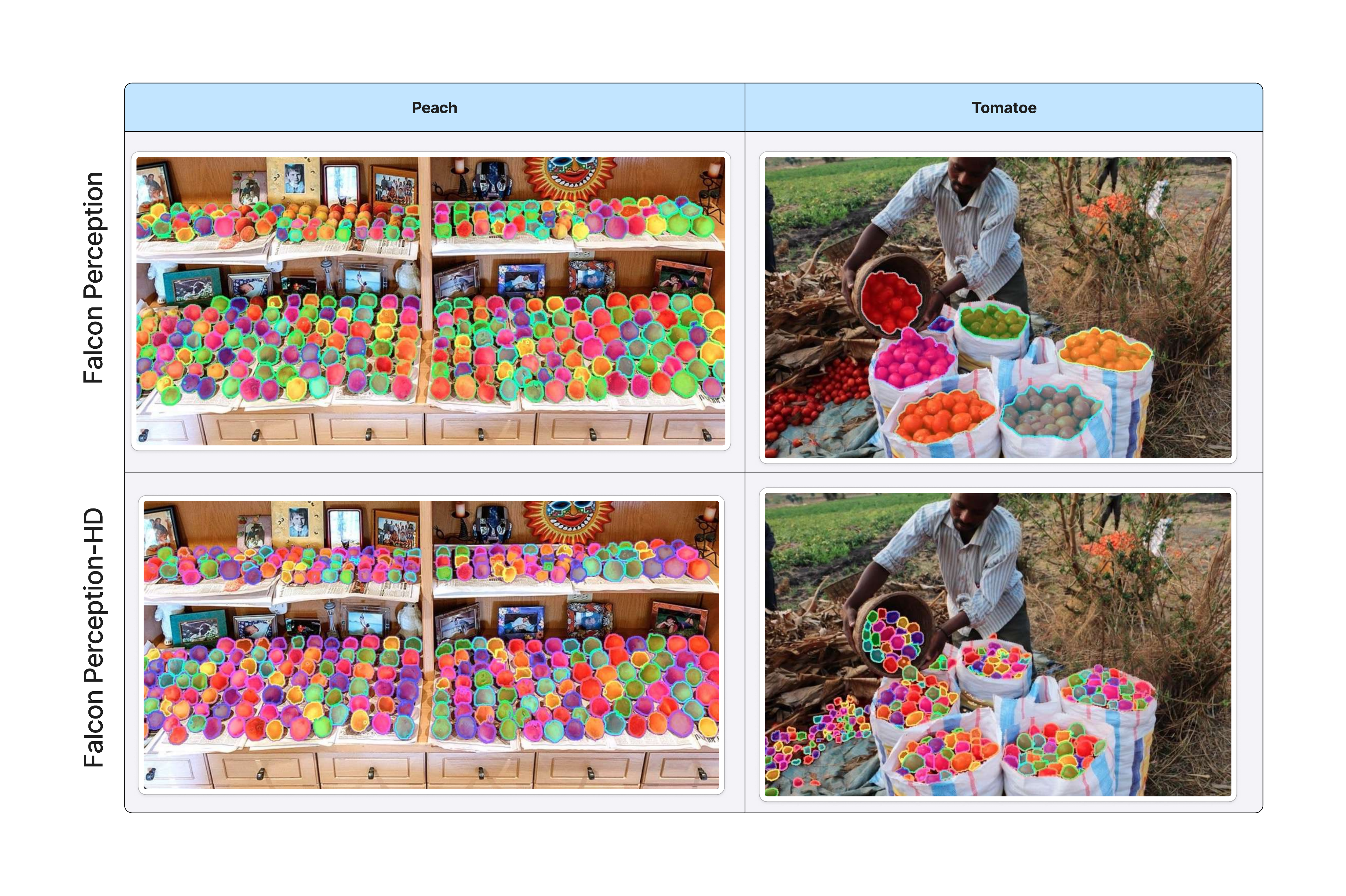}
    \caption{\textbf{Dense scenes, additional examples (2/2).} Top: Falcon Perception. Bottom: Falcon Perception-HD.}
    \label{fig:app_dense_2}
\end{figure}

We show two batches of additional dense-scene predictions in Figures~\ref{fig:app_dense_1} and~\ref{fig:app_dense_2}, complementing the teaser of Fig.~\ref{fig:teaser}. In each pair, the top row is the SFT-only baseline (Falcon Perception) and the bottom row is Falcon Perception-HD. The improvement is consistent: the post-trained model reaches recall on scenes containing several hundred instances of the queried category, where the baseline either drops a large fraction of objects or stops decoding prematurely.

\subsection{Level 4 (compositional referring expressions)}
\label{app:level4_qualitatives}

\begin{figure}[h]
    \centering
    \includegraphics[page=1, width=\textwidth]{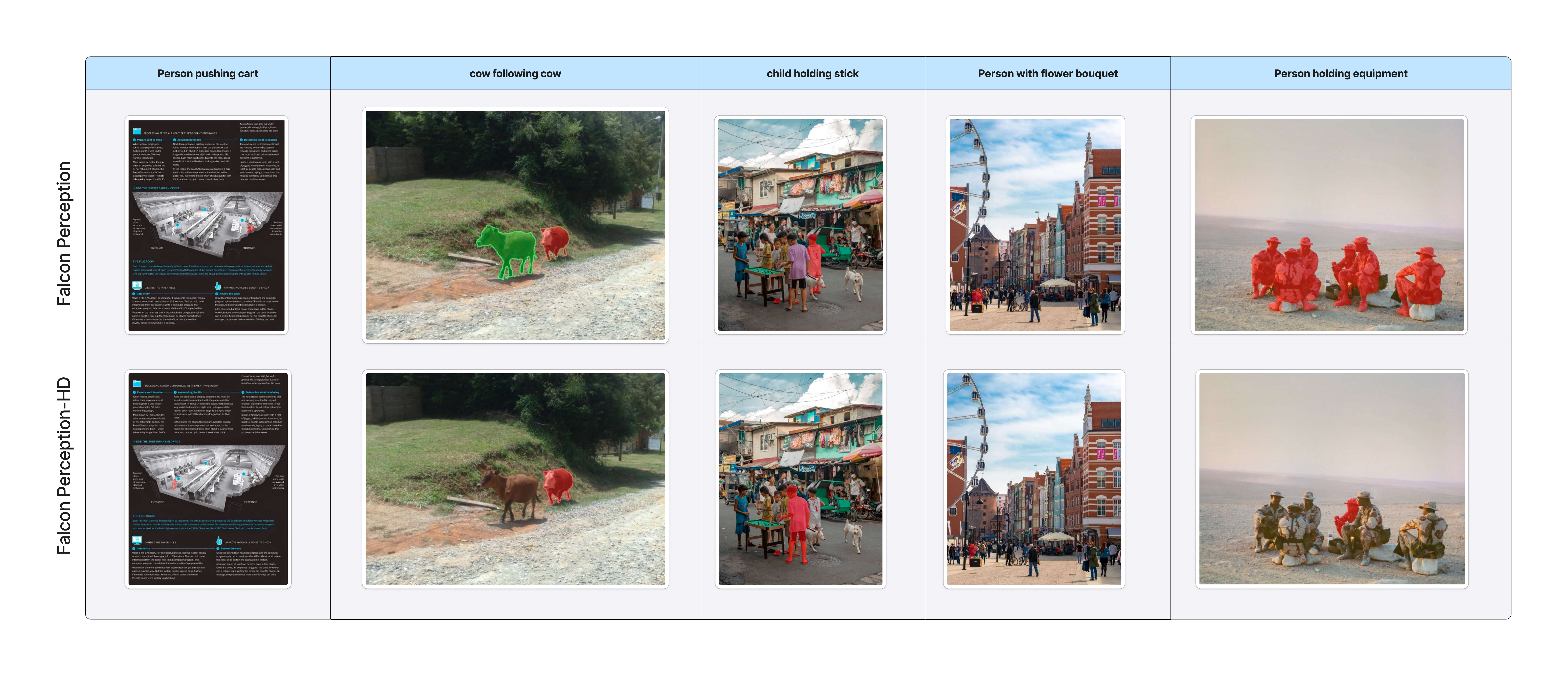}
    \caption{\textbf{Level 4 referring expressions, additional examples.} Top: Falcon Perception. Bottom: Falcon Perception-HD.}
    \label{fig:app_level4_more}
\end{figure}

Figure~\ref{fig:app_level4_more} shows additional level-4 examples illustrating compositional and relation-binding referring expressions, complementing Fig.~\ref{fig:level_4_qualitatives}.

\subsection{Mixed-difficulty PBench predictions}
\label{app:mix_qualitatives}

\begin{figure}[h]
    \centering
    \includegraphics[page=1, width=\textwidth]{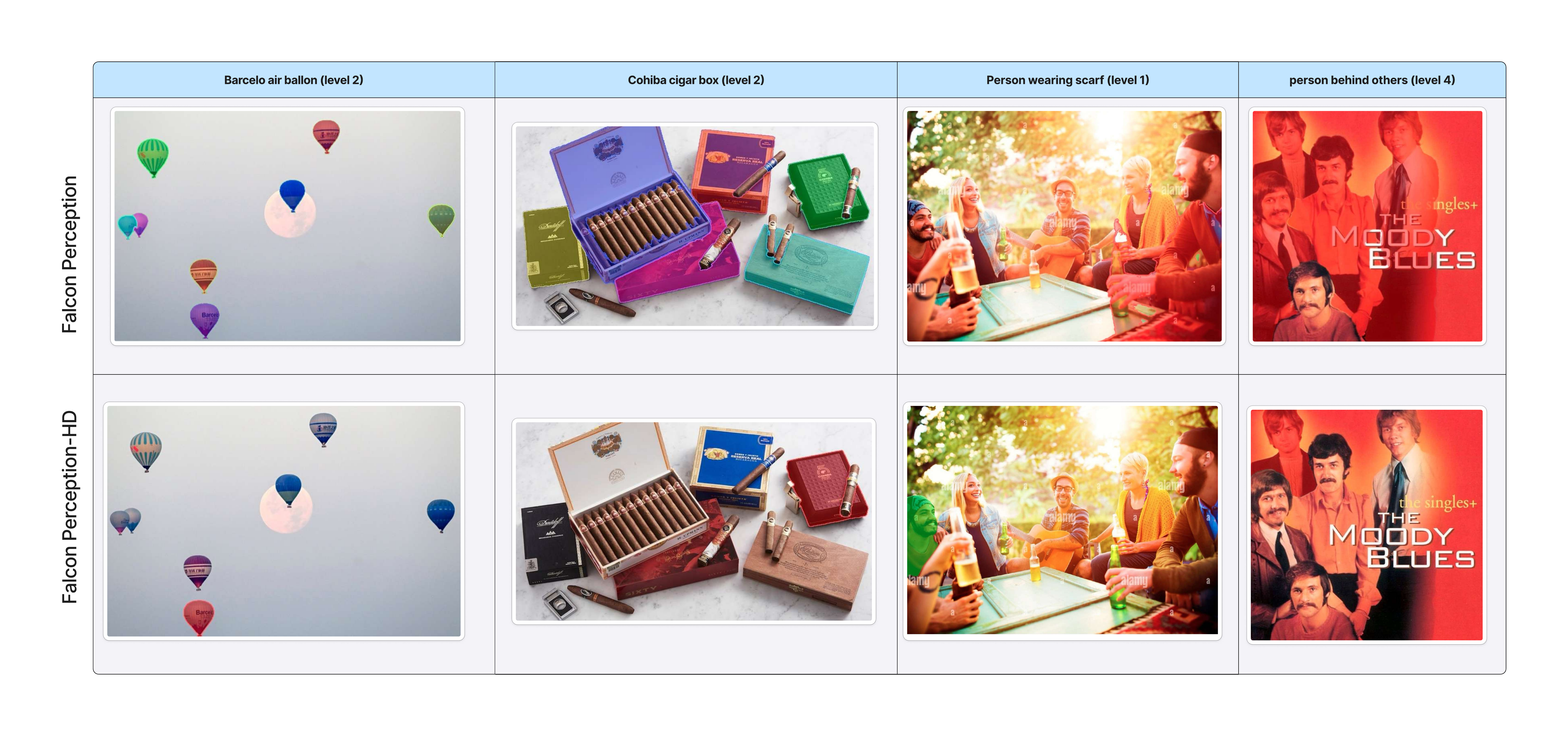}
    \caption{\textbf{Mixed PBench levels (non-dense), additional examples.} Top: Falcon Perception. Bottom: Falcon Perception-HD.}
    \label{fig:app_mix}
\end{figure}

Figure~\ref{fig:app_mix} shows examples sampled across PBench levels~0--4 (excluding the dense split). The post-trained model produces tighter boundaries and recovers objects missed by the baseline across difficulty levels, even though the segmentation head itself is frozen during RL: this is a direct manifestation of the cascade effect of \S\ref{sec:rl_objective}.

\subsection{Rollouts and the mechanism of RL}
\label{app:rollouts}

\begin{figure}[h]
    \centering
    \includegraphics[page=1, width=\textwidth]{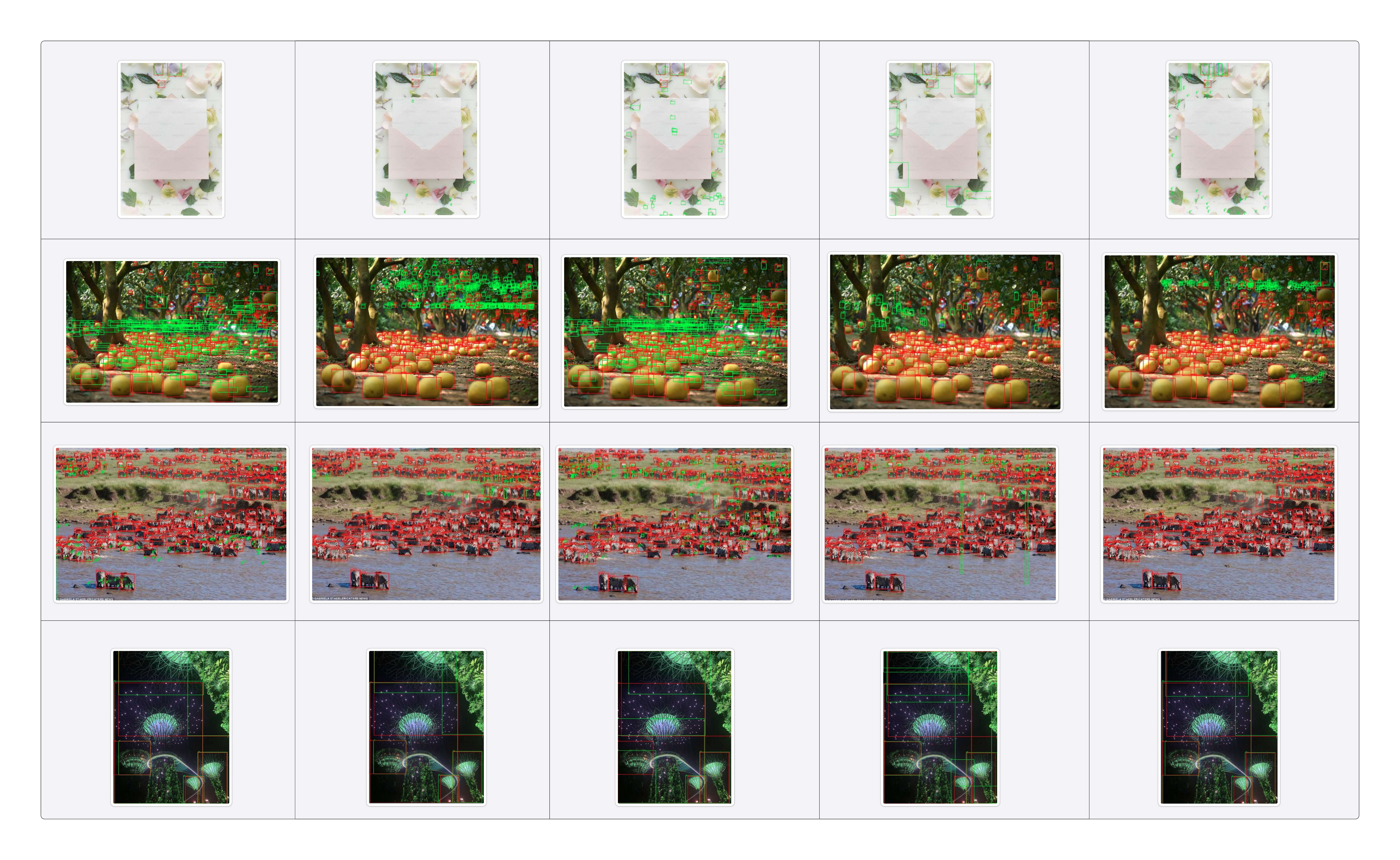}
    \caption{\textbf{Rollout samples from intermediate training checkpoints.} Each row shows several GRPO rollouts for the same prompt. Rollouts that emit near-duplicate predictions or that under-cover the scene receive negative within-group advantages and are suppressed by the policy gradient; this is the mechanism by which RL eliminates mask repetition (\S\ref{sec:nms_results}) and improves dense-scene recall (\S\ref{sec:dense_results}) without any explicit rule against either failure mode.}
    \label{fig:app_rollouts}
\end{figure}

Figure~\ref{fig:app_rollouts} provides a window into why RL works on autoregressive perception. For a small set of training prompts we display several rollouts sampled from the same intermediate policy. The rollouts show two recurring failure modes that GRPO naturally penalises:
\begin{itemize}
    \item \textbf{Indefinite or overlapping predictions.} Some rollouts emit near-duplicate detections at slightly perturbed coordinates, the policy-level analogue of the redundancy that NMS and coordinate deduplication are usually deployed to clean up after the fact (\S\ref{sec:nms_results}). Each duplicate is a Hungarian-matched false positive, so its rollout receives a strictly negative advantage relative to its siblings, and the policy gradient suppresses the corresponding token sequences.
    \item \textbf{Low-recall rollouts.} Other rollouts terminate early or skip large regions of the scene, scoring poorly on the count reward (Eq.~\ref{eq:count_reward}) compared to siblings that decode further. The negative advantage on these rollouts pushes the policy away from the early-EOS and regional-omission failure modes that drive the under-counting in dense scenes (\S\ref{sec:dense_results}).
\end{itemize}
Both effects are direct consequences of the group-relative advantage estimator: any failure mode that some siblings avoid will receive a within-group negative advantage and be suppressed, even though no rule explicitly forbids it.

\section{Self-annotation pipelines}
\label{app:annotation}

We illustrate here the final outputs of the two self-annotation pipelines described in \S\ref{sec:data_annotation}. We display only the post-pipeline annotations actually used for RL training (i.e., the human-verified outputs of the dense pipeline and the GPT-5-selected rollouts of the hard-refexp pipeline); intermediate stages of each pipeline are not shown.

\subsection{Dense scene annotations}
\label{app:dense_annot}

\begin{figure}[h]
    \centering
    \includegraphics[page=1, width=\textwidth]{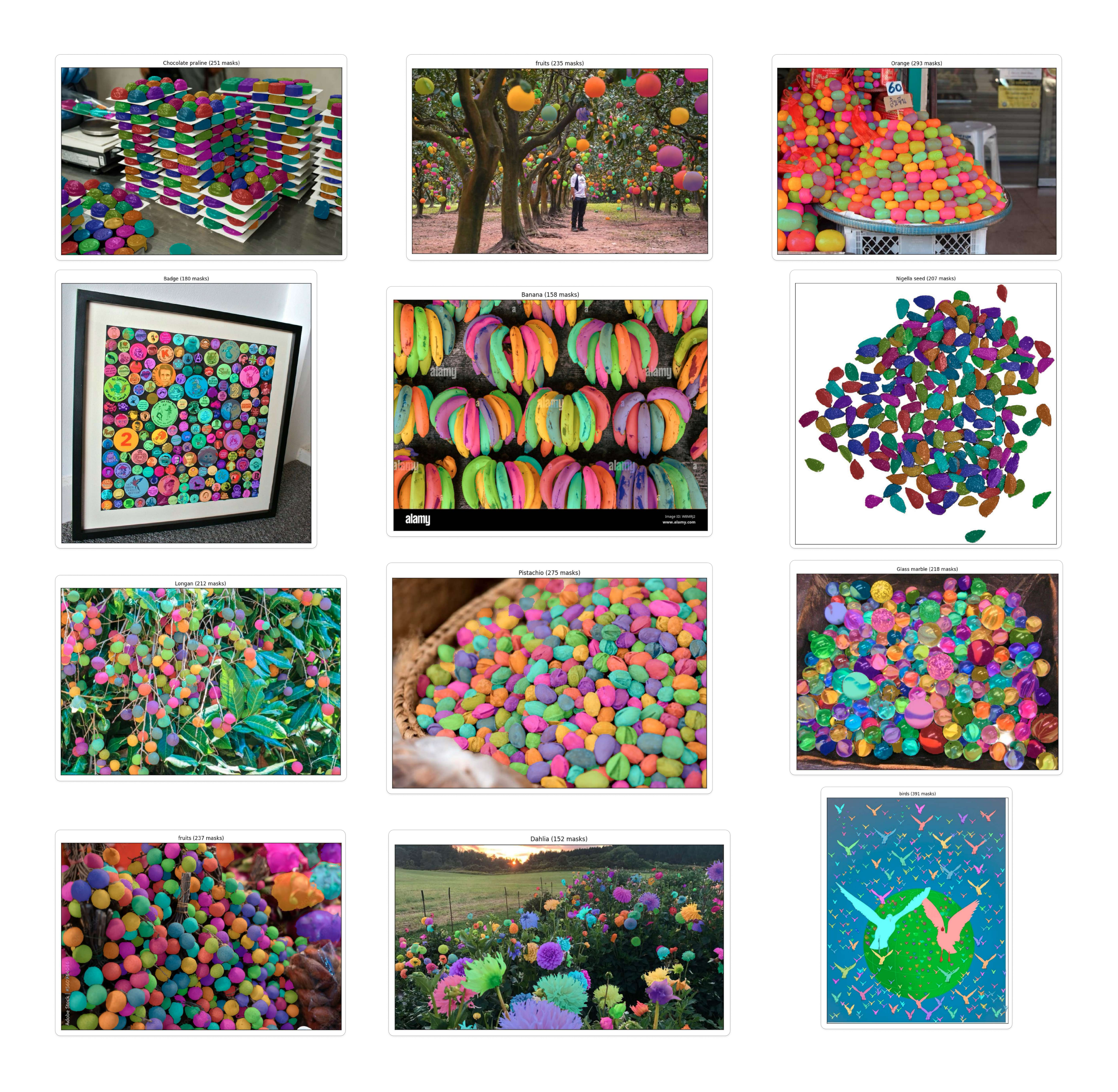}
    \caption{\textbf{Final dense-scene annotations.} Examples of the human-verified dense annotations used as ground truth during RL post-training.}
    \label{fig:app_dense_annot}
\end{figure}

Figure~\ref{fig:app_dense_annot} shows examples of the final annotations used for the dense split. Each image in this set contains up to several hundred instances of the queried category. The annotations are produced by running Falcon Perception over a curated set of dense images and then handing the predictions to human annotators for correction (removal of duplicate or fragmented masks, addition of missed instances, splitting of merged instances). The final masks form the supervised target on which the count reward of Eq.~\ref{eq:count_reward} is evaluated during RL.

\subsection{Hard referring-expression annotations}
\label{app:hard_refexp_annot}

\begin{figure}[h]
    \centering
    \includegraphics[page=1, width=\textwidth]{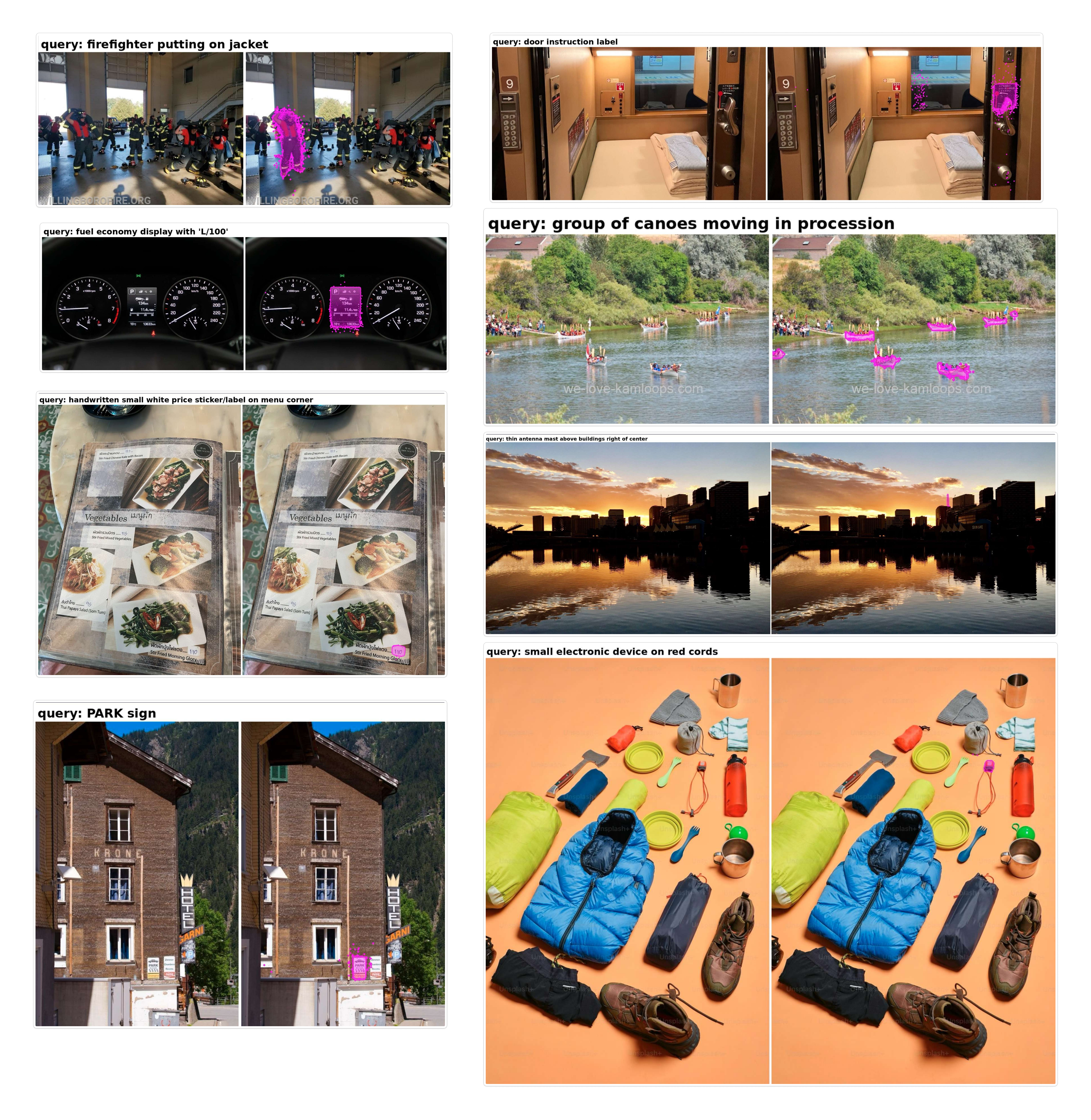}
    \caption{\textbf{Final hard-referring-expression annotations.} Each example shows the rollout selected by the GPT-5 judge as the ground-truth target for a prompt where the greedy prediction of Falcon Perception failed.}
    \label{fig:app_hard_refexp_annot}
\end{figure}

Figure~\ref{fig:app_hard_refexp_annot} shows examples of the final annotations used for the hard referring-expression set. Candidate annotations are produced by running pass@8 sampling on Falcon Perception over 200k image-expression pairs and keeping only those prompts where the greedy prediction is wrong but at least one sampled rollout is correct. The retained rollout, scored by GPT-5 as the most plausible candidate, is then used as the ground-truth mask for that prompt during RL. This restricts training to prompts that carry a useful policy-gradient signal (the greedy mode is wrong) while remaining within the support of the current policy (some rollout is right).

\paragraph{GPT-5 selection prompt.}
The exact prompt used to score each (image, query, deterministic prediction, $\{$rollout$\}$) tuple is reproduced verbatim below. The judge is shown the raw image, the temperature-0 greedy prediction, and $N$ stochastic rollouts (each with a magenta-overlay mask), and applies three criteria: (A) the greedy prediction is wrong, (B) at least one rollout is clearly correct, (C) there is meaningful intra-group variance. Only tuples that satisfy all three are kept.

\begin{tcolorbox}[breakable, enhanced, colback=gray!5, colframe=gray!50, boxrule=0.4pt, left=2mm, right=2mm, top=1mm, bottom=1mm, fontupper=\footnotesize\ttfamily]
You are helping curate data for reinforcement-learning training of a referring-segmentation model.\\[0.5em]
\textbf{\# What you are looking at}\\
You will receive, in this exact order:\\
1. \texttt{original} -- the raw input image with no overlay. Use this to understand the scene and judge what the query is actually asking about.\\
2. \texttt{deterministic} -- the model's greedy (temperature=0) prediction. The selected pixels are highlighted with a translucent \textbf{magenta fill} and a solid magenta border. The magenta color is an arbitrary annotation we add -- it is \textbf{not} a property of any object.\\
3. \texttt{rollout\_00}, \texttt{rollout\_01}, \ldots -- N stochastic samples from the same model for the same image+query, drawn the same way (magenta fill + magenta border).\\[0.3em]
The referring query and other metadata are provided as JSON in the next message (field \texttt{expression}).\\[0.5em]
\textbf{\# Your task}\\
For the given query, decide whether this sample is a good RL training example.\\[0.3em]
Evaluate each image independently:\\
$\bullet$ For \texttt{deterministic} and every \texttt{rollout\_XX}, look at the magenta-highlighted region and decide whether it correctly corresponds to the query. Note any wrong selections (wrong object, wrong part, spurious, missing). Give a short verdict: \texttt{correct}, \texttt{partially correct}, or \texttt{wrong}.\\[0.3em]
Then apply the selection criteria (ALL must hold to keep the sample):\\
(A) The \texttt{deterministic} prediction is wrong (or at best partially correct -- not a clean solution).\\
(B) At least one \texttt{rollout\_XX} is clearly correct.\\
(C) There is meaningful intra-group variance across rollouts -- they are not all highlighting the same region.\\[0.5em]
\textbf{\# Output format}\\
Respond in this exact structure:\\[0.3em]
\#\# Query\\
\textless restate the query\textgreater\\[0.3em]
\#\# Per-image judgment\\
- deterministic: \textless correct$|$partially correct$|$wrong\textgreater\ -- \textless one-line reason\textgreater\\
- rollout\_00: \textless correct$|$partially correct$|$wrong\textgreater\ -- \textless one-line reason\textgreater\\
- rollout\_01: \ldots\\
(one line per image)\\[0.3em]
\#\# Criteria check\\
- (A) deterministic wrong: \textless yes$|$no\textgreater\ -- \textless why\textgreater\\
- (B) at least one rollout correct: \textless yes$|$no\textgreater\ -- \textless which ones\textgreater\\
- (C) intra-rollout variance: \textless yes$|$no\textgreater\ -- \textless why\textgreater\\[0.3em]
\#\# Decision\\
KEEP or REJECT -- \textless one sentence justification\textgreater
\end{tcolorbox}

\section{Implementation Details}
\label{app:impl}

\subsection{Optimizer and schedule}
We use the Muon optimizer for all $\geq$2D weight tensors, with
Lion for 1D parameters (biases, RMSNorms), embeddings, and the
LM/coord/size heads. Hyperparameters: learning rate $5{\cdot}10^{-6}$,
$\beta_1{=}0.9, \beta_2{=}0.999$, $\varepsilon{=}10^{-8}$, weight decay
$0.01$, Muon momentum $\mu{=}0.95$, RMS-norm-based LR adjustment, cautious
weight decay, head LR scaled by $1/\sqrt{d}$. The schedule is a 50-step
linear warmup followed by linear decay over $80\%$ of training to a floor
of $0.1\times$ peak LR. The gradient is clipped at global norm $1.0$.

\subsection{Distributed setup and precision}
Training runs on $64$ GPUs with data-parallel replication and bf16 compute on top of the
torchtitan runtime. The local per-GPU batch size is $1$ prompt; with the group size $G{=}8$ used by GRPO, this expands to $8$ rollouts per GPU per
optimizer step, i.e. $512$ rollouts globally. Each prompt has $6{,}144$ maximum context (with up to $1024{\times}1024$
images) and we generate at most \texttt{max\_new\_tokens}${=}1856$ new tokens per rollout. We implement and use group packing: all sequences within the same group are packed together and share the same prompt (Image+query). This is possible thanks to FlexAttention, which allows masking out self-attention between tokens from different rollouts. 

\subsection{Rollout engine}
\label{app:impl:rollout}
Rollouts are produced by a custom paged-attention inference engine built on FlexAttention.

\section{Additional ablations}
\label{app:ablations}

\subsection{Loss aggregation: Dr.\ GRPO vs.\ standard GRPO}
\label{app:drgrpo_ablation}

We argue in \S\ref{sec:rl_objective} that the standard GRPO loss aggregation, which divides each rollout's contribution by its own length $1/T_i$, biases the policy towards shorter responses, and that the Dr.\ GRPO denominator $1/(B \cdot T_\text{max})$ removes this bias. Table~\ref{tab:app_drgrpo} verifies this on the high-density buckets of PBench, where the bias is expected to bite hardest because the desired rollouts can exceed $3{\cdot}10^4$ tokens. We compare two runs trained with the same recipe but different aggregation, both selected at their best checkpoint by overall average F$_1$, and report F$_1$ on the four highest-density buckets ($\geq 50$ instances per scene). Both schemes improve substantially over the pre-RL baseline on the moderate-density buckets, and standard GRPO is in fact slightly ahead on $50{-}300$ objects per scene; the picture inverts on the most extreme density bucket ($300{+}$), where Dr.\ GRPO dominates by $+8.5$ F$_1$. This is consistent with the analysis: standard GRPO's per-token weighting under-counts the very long rollouts needed for hyper-dense scenes, so the policy never gets pushed to sustain $1000$-token decoding to completion. Dr.\ GRPO, whose per-token gradient is independent of $T_i$, retains the high-density gain.

\begin{table}[h]
    \centering
    \small
    \caption{\textbf{Loss aggregation ablation, PBench segmentation F$_1$ on high-density buckets ($\geq 50$ instances per scene).} Standard GRPO performs comparably on moderate density but loses ground in the extreme-density regime ($300+$), where its per-token weighting under-counts very long rollouts. Dr.\ GRPO removes this length bias and retains the high-density gain.}
    \label{tab:app_drgrpo}
    \setlength{\tabcolsep}{6pt}
    \renewcommand{\arraystretch}{1.15}
    \adjustbox{max width=\textwidth}{
    \begin{tabular}{lcccc|c}
        \toprule
        \rowcolor{headergray}
        \textbf{Aggregation} & \textbf{50--100} & \textbf{100--200} & \textbf{200--300} & \textbf{300+} & \textbf{Avg ($\geq 50$)} \\
        \midrule
        Falcon Perception (pre-RL)            & 78.7 & 76.5 & 67.4 & 55.9 & 69.6 \\
        \quad + RL, standard GRPO ($1/T_i$)   & 80.6 & \textbf{81.4} & \textbf{78.7} & 61.2 & 75.5 \\
        \rowcolor{ourrow}
        \quad + RL, Dr.\ GRPO \textbf{(ours)} & \textbf{80.2} & 80.4 & 74.5 & \textbf{69.8} & \textbf{76.2} \\
        \bottomrule
    \end{tabular}
    }
\end{table}

\subsection{Clip-Cov on/off}
\label{app:clipcov_ablation}

Table~\ref{tab:app_clipcov} compares the main recipe (Clip-Cov enabled, with the per-head bounds reported in \S\ref{sec:rl_objective}) against an otherwise identical run with Clip-Cov disabled. We report the best checkpoint of each run on PBench, selected by average F$_1$ across the six splits; both runs peak at step~550 of training.

Clip-Cov yields a small but consistent improvement on every PBench level except L0, with the largest absolute gain on L2 ($+0.5$ F$_1$). This is consistent with the hypothesis that Clip-Cov mitigates over-reinforcement of confidently-wrong rollouts: L2 (OCR in natural scenes) is the level on which the reward signal is most contaminated by annotation noise, since dense-text scenes have boundary and identity ambiguities that propagate directly into the count reward. By zeroing the gradient contribution of tokens with the highest log-probability/advantage covariance, Clip-Cov prevents the policy from sharpening on rollouts whose high reward comes from a noisy ground truth rather than from a correct prediction.

\begin{table}[h]
    \centering
    \small
    \caption{\textbf{Clip-Cov on/off, PBench segmentation F$_1$.} Disabling Clip-Cov costs $0.3$ F$_1$ on average and is most visible on L2 (OCR in natural scenes), where annotation noise dominates and over-reinforcement of confidently-wrong rollouts is most costly.}
    \label{tab:app_clipcov}
    \setlength{\tabcolsep}{6pt}
    \renewcommand{\arraystretch}{1.15}
    \adjustbox{max width=\textwidth}{
    \begin{tabular}{lccccccc|c}
        \toprule
        \rowcolor{headergray}
        \textbf{Setting} & \textbf{L0} & \textbf{L1} & \textbf{L2} & \textbf{L3} & \textbf{L4} & \textbf{Dense} & & \textbf{Avg} \\
        \midrule
        Falcon Perception (pre-RL)    & 63.7 & 63.8 & 38.3 & 53.4 & 49.1 & 72.3 & & 56.8 \\
        \quad + RL, no Clip-Cov        & 64.9 & 63.8 & 39.9 & 54.7 & 51.4 & 80.1 & & 59.1 \\
        \rowcolor{ourrow}
        \quad + RL, with Clip-Cov \textbf{(ours)} & \textbf{64.9} & \textbf{64.2} & \textbf{40.4} & \textbf{54.7} & \textbf{51.8} & \textbf{80.5} & & \textbf{59.4} \\
        \midrule
        \rowcolor{deltarow}
        $\Delta$ Clip-Cov &
        \textcolor{ForestGreen}{$+0.0$} &
        \textcolor{ForestGreen}{$+0.4$} &
        \textcolor{ForestGreen}{\textbf{$+0.5$}} &
        \textcolor{ForestGreen}{$+0.0$} &
        \textcolor{ForestGreen}{$+0.4$} &
        \textcolor{ForestGreen}{$+0.4$} & &
        \textcolor{ForestGreen}{\textbf{$+0.3$}} \\
        \bottomrule
    \end{tabular}
    }
\end{table}

\subsection{Finer-grained rewards and locality metrics}
\label{app:locality_ablation}

Our count reward scores detection only, so one may ask whether a finer-grained reward that scores localization quality improves locality metrics. We train an otherwise identical model with the panoptic reward of \citet{pinto2023tuning}, which favors better IoU, and report mean matched-mask IoU and macro-F$_1$ per PBench level in Table~\ref{tab:app_locality}. We see no significant difference between the two rewards: both improve over the pre-RL baseline, on locality metrics as well. This is a consequence of the chain-of-perception and the cascade effect, where good pointing implies good segmentation, so the simpler count reward is sufficient.

\begin{table}[h]
    \centering
    \small
    \caption{\textbf{Count vs.\ panoptic reward, PBench.} Mean matched-mask IoU and macro-F$_1$ per level. The finer-grained panoptic reward does not improve locality metrics over the count reward: through the cascade effect, good pointing already implies good segmentation.}
    \label{tab:app_locality}
    \setlength{\tabcolsep}{6pt}
    \renewcommand{\arraystretch}{1.15}
    \adjustbox{max width=\textwidth}{
    \begin{tabular}{lccccccc}
        \toprule
        \rowcolor{headergray}
        \textbf{Reward} & \textbf{L0} & \textbf{L1} & \textbf{L2} & \textbf{L3} & \textbf{L4} & \textbf{Dense} & \textbf{Avg} \\
        \midrule
        \multicolumn{8}{l}{\emph{Mean matched-mask IoU}} \\
        pre-RL baseline & 78.2 & 77.0 & 52.0 & 63.9 & 64.6 & 68.7 & 67.4 \\
        count           & \textbf{79.1} & \textbf{76.5} & \textbf{51.2} & \textbf{64.1} & \textbf{64.9} & \textbf{72.4} & \textbf{68.0} \\
        panoptic        & 79.0 & 76.0 & 50.9 & 63.8 & 64.6 & 72.0 & 67.7 \\
        \midrule
        \multicolumn{8}{l}{\emph{Macro-F$_1$}} \\
        pre-RL baseline & 63.8 & 63.8 & 38.3 & 53.4 & 49.1 & 71.9 & 56.7 \\
        count           & 64.2 & 63.2 & 39.4 & 54.2 & 50.5 & \textbf{80.9} & \textbf{58.7} \\
        panoptic        & \textbf{64.3} & 63.1 & \textbf{39.5} & \textbf{54.3} & \textbf{50.6} & 80.2 & \textbf{58.7} \\
        \bottomrule
    \end{tabular}
    }
\end{table}

\section{Extension to other autoregressive models: Pix2Seq}
\label{app:pix2seq}

We verify that our two central conclusions, dense perception and the removal of NMS, i.e., RL serving as a surrogate for Hungarian matching in autoregressive models, generalize to a standard autoregressive model independent of Falcon Perception and of our data pipeline. We take the pioneering work in this area, Pix2Seq~\cite{pix2seq}, and build a simple 70M-parameter model (a vision encoder and a 6-layer autoregressive decoder) trained from scratch for closed-vocabulary detection on the open-source VisDrone and COCO datasets. The task is to list and detect all the objects present in the image in the format $[y_{\min}, x_{\min}, y_{\max}, x_{\max}, \text{class}, \dots]$; there is no detection-specific head, and positions are encoded as $500$ discrete bins that are part of the full vocabulary (classes and positions). We train the model with maximum likelihood for ${\sim}50$k steps at batch size $128$, then RL post-train it for $500$ steps with our Hungarian count reward aggregated over all objects in the image, irrespective of the class.

Table~\ref{tab:app_pix2seq} reports F$_1$@0.5 on the validation sets, with and without NMS. The conclusions are the same as for Falcon Perception: SFT does not fix the overprediction of boxes (removing NMS costs the base model $23.1$ and $10.9$ points), while RL alone, with a much lower budget, entirely removes the need for NMS and translates into large performance improvements. Qualitatively, the base model, while properly detecting existing objects, repeats itself, predicts degenerate boxes at random places in the image, and does not know how to terminate the sequence (the EOS token is never emitted). RL quickly fixes all three behaviors by penalizing them during training. Pix2Seq reported the same termination issue and addressed it by preventing EOS emission and learning a noise token~\cite{pix2seq}; a fundamental limitation of this heuristic is that the token budget does not adapt to scene density. We conclude that RL post-training impacts autoregressive perception models the way Hungarian matching and DETR impacted classical detection models: it removes the need for these heuristics entirely.

\begin{table}[h]
    \centering
    \small
    \caption{\textbf{Pix2Seq-style model, F$_1$@0.5 on the VisDrone and COCO validation sets.} $500$ steps of RL entirely remove the raw$\to$NMS gap and yield large absolute gains, replicating our Falcon Perception findings on an independent architecture, data, and codebase.}
    \label{tab:app_pix2seq}
    \setlength{\tabcolsep}{6pt}
    \renewcommand{\arraystretch}{1.15}
    \adjustbox{max width=\textwidth}{
    \begin{tabular}{l ccc c ccc}
        \toprule
        \rowcolor{headergray}
        & \multicolumn{3}{c}{\textbf{VisDrone}} & & \multicolumn{3}{c}{\textbf{COCO}} \\
        \rowcolor{headergray}
        \textbf{Model} & raw & NMS & raw$\to$NMS gap & & raw & NMS & raw$\to$NMS gap \\
        \midrule
        Base (SFT) & 21.2 & 44.3 & \textcolor{red}{$+23.1$} & & 14.3 & 25.2 & \textcolor{red}{$+10.9$} \\
        \rowcolor{ourrow}
        \textbf{RL} & \textbf{60.6} & \textbf{60.8} & \textcolor{ForestGreen}{$\mathbf{+0.2}$} & & \textbf{59.6} & \textbf{59.8} & \textcolor{ForestGreen}{$\mathbf{+0.2}$} \\
        \bottomrule
    \end{tabular}
    }
\end{table}

\section{External dense benchmarks}
\label{app:external_bench}

To reinforce our claims beyond the benchmarks tied to the Falcon Perception ecosystem, we evaluate on external dense benchmarks. \emph{COCO-dense} uses COCO val2017 in a category-as-query setting restricted to queries with at least $5$ instances ($1{,}974$ queries); \emph{LVIS-dense} keeps the $882$ LVIS queries with more than $25$ instances; \emph{Dense200}~\cite{jiang2026detect} is a recently introduced dense detection benchmark (box F$_1$@0.5). We also evaluate two recent strong detection models, LocateAnything-3B~\cite{wang2026locateanything} and Rex-Omni-3B~\cite{jiang2026detect}, on the PBench dense split. These models emit boxes only, so this comparison uses box F$_1$ at IoU $0.5$, with ground-truth boxes derived from the PBench dense masks, applied identically to all models.

Tables~\ref{tab:app_external} and~\ref{tab:app_pbench_box} report the results. On every benchmark, the gains of the RL post-trained model over the baseline are considerable. On Dense200, Falcon Perception-HD is competitive with the best specialized models, Rex-Omni and LocateAnything (detection only), despite being $5\times$ smaller. In contrast, these models perform comparatively poorly on the PBench dense split and exhibit heavily redundant predictions ($28.5\%$ MRR for LocateAnything), again requiring NMS.

\begin{table}[h]
    \centering
    \small
    \caption{\textbf{External dense benchmarks.} Left: mask F$_1$ on COCO-dense (category-as-query, $\geq 5$ instances, $1{,}974$ queries) and LVIS-dense ($882$ queries, $>25$ instances). Right: box F$_1$@0.5 on Dense200.}
    \label{tab:app_external}
    \setlength{\tabcolsep}{6pt}
    \renewcommand{\arraystretch}{1.15}
    \begin{minipage}[t]{0.48\textwidth}
    \centering
    \begin{tabular}{lcc}
        \toprule
        \rowcolor{headergray}
        \textbf{Model} & \textbf{COCO-dense} & \textbf{LVIS-dense} \\
        \midrule
        SAM 3        & 43.5 & 31.5 \\
        Baseline FP  & 55.1 & 33.6 \\
        \rowcolor{ourrow}
        \textbf{FP-HD (Ours)} & \textbf{58.1} & \textbf{40.5} \\
        \bottomrule
    \end{tabular}
    \end{minipage}
    \hfill
    \begin{minipage}[t]{0.48\textwidth}
    \centering
    \begin{tabular}{lc}
        \toprule
        \rowcolor{headergray}
        \textbf{Model} & \textbf{Dense200 F$_1$@0.5} \\
        \midrule
        SAM 3               & 63.3 \\
        LocateAnything-3B   & 74.0 \\
        Rex-Omni-3B         & \textbf{78.4} \\
        Baseline FP         & 72.0 \\
        \rowcolor{ourrow}
        \textbf{FP-HD (Ours)} & \textit{78.1} \\
        \bottomrule
    \end{tabular}
    \end{minipage}
\end{table}

\begin{table}[h]
    \centering
    \small
    \caption{\textbf{Box F$_1$@0.5 on the PBench dense split.} Ground-truth boxes are derived from the PBench dense masks; the protocol is identical for all models. NMS yields only marginal improvements on the redundant baselines, while Falcon Perception-HD does not need it.}
    \label{tab:app_pbench_box}
    \setlength{\tabcolsep}{8pt}
    \renewcommand{\arraystretch}{1.15}
    \begin{tabular}{lccc}
        \toprule
        \rowcolor{headergray}
        \textbf{Model} & \textbf{no NMS} & \textbf{with NMS} & \textbf{MRR} \\
        \midrule
        LocateAnything-3B~\cite{wang2026locateanything} & 35.9 & 36.6 & \textcolor{red}{28.5\%} \\
        Rex-Omni-3B~\cite{jiang2026detect}              & ---  & 37.4 & --- \\
        Falcon Perception (pre-RL)              & 67.9 & 68.0 & 2.1\% \\
        \rowcolor{ourrow}
        \textbf{FP-HD (Ours)}                   & \textbf{73.9} & \textbf{73.9} & \textbf{0.4\%} \\
        \bottomrule
    \end{tabular}
\end{table}

\section{Autoregressive vs.\ DETR-style models}
\label{app:ar_vs_detr}

Our RL framework targets autoregressive models. DETR-style models like SAM 3 already solve the repetition problem natively during base training via bipartite matching, and we verify empirically that RL post-training is unlikely to help them. Since these models do not sample, we formulate a stochastic policy for SAM 3 as follows: a well-defined stochastic choice is which of its $200$ decoder queries to emit. Each query carries a per-slot probability $p_i = \sigma(\text{cls}_i)$, and we treat the selection as a product of independent Bernoullis, $a \sim \prod_i \mathrm{Bernoulli}(p_i)$, $a \in \{0,1\}^{200}$. Table~\ref{tab:app_passk} reports pass@$k$ on the PBench dense split under this policy: pass@$8$ is worse than the deterministic prediction ($-3.4$), while Falcon Perception gains $+6.9$. The core premise of RL post-training, that some rollouts beat the deterministic prediction, does not hold for SAM 3.

This native robustness comes at a cost: DETR models are strictly limited by their fixed number of object queries (SAM 3 is capped at $200$), whereas autoregressive models can dynamically generate as many objects as required; Falcon Perception-HD successfully detects up to $600$ objects in a single generation. We further benchmarked widely used autoregressive VLMs, Moondream3~\cite{korrapati2024moondream} and Qwen3.5 (4B/8B)~\cite{qwen2025qwen3vl}, on the PBench dense split and found their box redundancy rate exceeds $30\%$, also requiring NMS. This suggests that duplicate prediction is an algorithmic flaw of token-level MLE training, not a data issue, and a short RL stage fixes it entirely. We conclude that RL post-training for autoregressive perception models acts as the equivalent of DETR's Hungarian matching.

\begin{table}[h]
    \centering
    \small
    \caption{\textbf{pass@$k$ on the PBench dense split.} Stochastic sampling improves over the deterministic prediction for the autoregressive model, which motivates RL post-training, but degrades the DETR-style SAM 3 under the per-query Bernoulli policy.}
    \label{tab:app_passk}
    \setlength{\tabcolsep}{8pt}
    \renewcommand{\arraystretch}{1.15}
    \begin{tabular}{lcc}
        \toprule
        \rowcolor{headergray}
        $k$ & \textbf{Falcon Perception (AR)} & \textbf{SAM 3 (DETR)} \\
        \midrule
        deterministic & 72.1 & 58.4 \\
        pass@2        & 72.7 & 54.5 \\
        pass@4        & 76.8 & 55.7 \\
        pass@8        & \textbf{79.0} & 56.8 \\
        \midrule
        \rowcolor{deltarow}
        $\Delta$ (pass@8 $-$ deterministic) & \textcolor{ForestGreen}{$\mathbf{+6.9}$} & \textcolor{red}{$\mathbf{-3.4}$} \\
        \bottomrule
    \end{tabular}
\end{table}

\section{Training curves}
\label{app:wandb}

\begin{figure}[H]
    \centering
    \includegraphics[width=\textwidth]{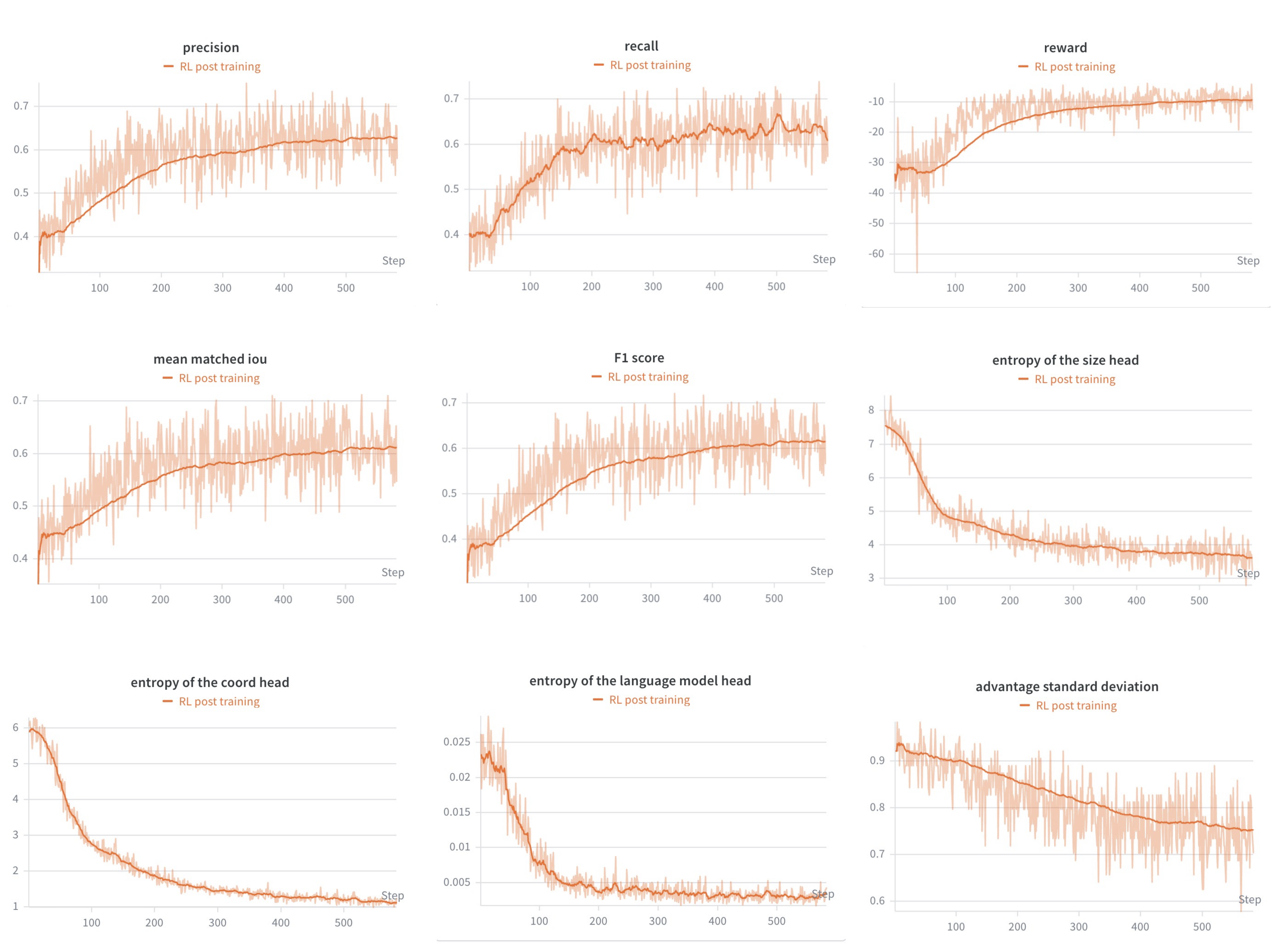}
    \caption{\textbf{Training curves.} Top: total reward, precision, recall, mean matched IoU, and F$_1$. Bottom: per-head policy entropy (LM, coordinate, size) and within-group advantage standard deviation. The size head is frozen and sampled greedily, yet its entropy decreases in lockstep with the two trainable heads, an in-training manifestation of the cascade effect of \S\ref{sec:rl_objective}.}
    \label{fig:app_training_curves}
\end{figure}

Figure~\ref{fig:app_training_curves} reports the training-time dynamics of our main run over the full 600-step horizon. The top row tracks the reward signals: total reward, mean precision and recall under Hungarian matching, mean matched IoU, and the resulting F$_1$ all move monotonically and saturate by step $\sim$550, which is the checkpoint reported in the main results. The bottom row reports per-head policy entropies (LM, coordinate, size) and the standard deviation of within-group advantages. The LM and coordinate heads, which are the only heads on which the policy gradient is propagated, see their entropy decrease as the policy concentrates on high-reward modes; the advantage standard deviation decays accordingly, indicating that within-group disagreement shrinks as the rollouts of the same prompt converge towards similar high-reward layouts.

Crucially, the size-head entropy, although the size head is held frozen and sampled at $\tau{=}0$ throughout training, follows the same decreasing trajectory as the two heads that do receive gradients. Since no gradient ever flows through the size head, the only mechanism by which its entropy can change is the cascade effect of the chain-of-perception decoder: improving the LM and coordinate predictions sharpens the hidden-state input to the size head, which in turn reduces its sampling entropy. This is the in-training counterpart of the cascade effect we identify in \S\ref{sec:rl_objective} and quantify in the main tables, observed here as a clean signature on a head that is never directly optimized.

\section{Mathematical Derivations}

This appendix section contains the three derivations that justify non-obvious design choices of \S\ref{sec:method}: (i) why the detached importance-sampling ratio of Eq.~\ref{eq:is_weight} yields an unbiased on-policy estimator under engine mismatch (\S \ref{app:E1}); (ii) why standard GRPO loss aggregation systematically biases the policy toward shorter rollouts and how Dr.~GRPO (Eq.~(\ref{eq:dr_grpo}))
removes the bias (\S \ref{app:E2}); and (iii) the explicit gradient bias on existence tokens under positive-only RL that motivates the stop-gradient of \S3.4 (\S \ref{app:E3}).

\paragraph{Notation.}
A rollout is $o = (a_1, \dots, a_T) \sim \pi_\theta(\cdot \mid I, q)$ with state $s_t$ at step $t$ and reward $r(o)$ from Eq.~\ref{eq:count_reward}. Group-relative advantage $\hat{A}_i = (r_i - \mu_G)/(\sigma_G + \epsilon)$ is computed over $G$ rollouts from the same prompt. We write $\theta_{\mathrm{rollout}}$ for the parameter snapshot used by the inference engine and $\theta$ for the parameter at gradient-computation time; in single-update GRPO the two are bit-identical, but the engines differ.

\subsection{Engine-mismatch importance sampling}
\label{app:E1}

Let $\tilde{p}(a \mid s) := \pi_{\theta_{\mathrm{rollout}}}(a \mid s)$ be the distribution sampled by the inference engine and $p(a \mid s) :=\pi_\theta(a \mid s)$ the distribution evaluated by the training engine.
Even at identical parameters $\theta_{\mathrm{rollout}} = \theta$, the two engines use different attention kernels and accumulate floating-point error differently, so $\tilde{p} \ne p$ in general. We want the on-policy gradient under $p$:
\begin{equation*}
\nabla_\theta J(\theta) \;=\;
\mathbb{E}_{o \sim p}\!\left[r(o) \sum_t \nabla_\theta \log p(a_t \mid s_t)\right],
\end{equation*}
but our samples come from $\tilde{p}$. Importance sampling rewrites this as an expectation under $\tilde{p}$, with per-token ratio $w_t := p(a_t \mid s_t) / \tilde{p}(a_t \mid s_t)$. The VeRL implementation we follow applies the per-token ratio multiplicatively with stop-gradient:
\begin{equation}
\ell_t^{(i)}
\;=\;
- \mathrm{sg}\!\left(w_t^{(i)}\right) \cdot \hat{A}_i \cdot \log p\!\left(a_t^{(i)} \mid s_t^{(i)}\right),
\label{eq:app-is-loss}
\end{equation}
which corresponds to Eq.~\ref{eq:loss_is}.
 
\paragraph{Unbiasedness with stop-gradient.}
Under stop-gradient on $w_t$, the per-token loss
\eqref{eq:app-is-loss} satisfies
$- \mathbb{E}_{o \sim \tilde{p}}\!\left[\nabla_\theta \ell_t^{(i)}\right]
=
\mathbb{E}_{o \sim p}\!\left[\hat{A}_i \cdot \nabla_\theta \log p(a_t^{(i)} \mid s_t^{(i)})\right]$,
i.e., the gradient estimator is unbiased for the on-policy gradient
under $p$. To see this, note that stop-gradient makes $w_t$ a constant w.r.t.\ $\theta$, so $\nabla_\theta \ell_t^{(i)} = - w_t^{(i)} \cdot \hat{A}_i \cdot \nabla_\theta \log p(a_t^{(i)} \mid s_t^{(i)})$. Taking expectation under $\tilde{p}$ and using $w_t = p / \tilde{p}$,
\begin{equation*}
\mathbb{E}_{o \sim \tilde{p}}\!\left[
\tfrac{p(a_t \mid s_t)}{\tilde{p}(a_t \mid s_t)} \cdot
\hat{A} \cdot \nabla_\theta \log p(a_t \mid s_t)
\right]
\;=\;
\mathbb{E}_{o \sim p}\!\left[
\hat{A} \cdot \nabla_\theta \log p(a_t \mid s_t)
\right],
\end{equation*}
which is the on-policy gradient.
 
\paragraph{Why the stop-gradient is necessary.}
Without stop-gradient, $w_t = p(a_t \mid s_t)/\tilde{p}(a_t \mid s_t)$ would also depend on $\theta$ (since $\tilde{p}$ is held fixed at the rollout-time snapshot). The product $w_t \log p(a_t \mid s_t)$ would then contribute two score-function-like terms when differentiated, double-counting the gradient and producing a biased estimator. Stop-gradient on $w_t$ restores the standard importance-sampling identity.
 
\paragraph{Why clipping has no effect in our setting.}
PPO-style clipping of $w_t$ acts as a trust-region mechanism for cases where $\theta_{\mathrm{rollout}} \ne \theta$, e.g., multiple gradient steps per rollout group. We perform a single update per group, so the only source of $w_t \ne 1$ is engine-level numerical noise; in practice $w_t$ concentrates tightly around $1$ and the clip range is never active.
The IS correction therefore acts as a numerical-stability fix rather than as a trust region.

\subsection{Length bias in standard GRPO loss aggregation}
\label{app:E2}
Standard GRPO aggregates per-token losses by averaging within each
rollout and then across rollouts:
\begin{equation}
\mathcal{L}_{\mathrm{std}}(\theta)
\;=\;
\mathbb{E}_{q,I}\!\left[
\frac{1}{G}\sum_{i=1}^{G} \frac{1}{T_i}
\sum_{t=1}^{T_i} \ell_t^{(i)}
\right].
\label{eq:app-loss-std}
\end{equation}
Dr.~GRPO replaces the per-rollout factor $1/T_i$ with a fixed constant
$1/(B \cdot T_{\max})$ that does not depend on the realized length:
\begin{equation}
\mathcal{L}_{\mathrm{drGRPO}}(\theta)
\;=\;
\mathbb{E}_{q,I}\!\left[
\frac{1}{B \cdot T_{\max}}
\sum_{i=1}^{G} \sum_{t=1}^{T_i} \ell_t^{(i)}
\right].
\label{eq:app-loss-drgrpo}
\end{equation}
 
\paragraph{Per-token gradient mismatch under $\mathcal{L}_{\mathrm{std}}$.}
Consider two rollouts $o_i, o_j$ in the same group with the same
advantage $\hat{A}_i = \hat{A}_j = \hat{A}$ and the same per-token score function $g$ (in expectation). The contribution of a single token to the gradient of $\mathcal{L}_{\mathrm{std}}$ is
\begin{equation*}
\frac{\partial \mathcal{L}_{\mathrm{std}}}{\partial \theta}
\;\supset\;
\frac{1}{G \cdot T_i}\,w_t \,\hat{A}\, g,
\qquad\text{vs.}\qquad
\frac{\partial \mathcal{L}_{\mathrm{drGRPO}}}{\partial \theta}
\;\supset\;
\frac{1}{B \cdot T_{\max}}\,w_t\,\hat{A}\, g,
\end{equation*}
so a token in $o_i$ receives a per-token gradient larger by a factor $T_j / T_i$ relative to a token in $o_j$ under
$\mathcal{L}_{\mathrm{std}}$, while both receive equal per-token gradient under $\mathcal{L}_{\mathrm{drGRPO}}$.
 
\paragraph{Direction of the bias in our setting.}
In dense detection, a high-reward rollout that successfully detects
$\sim 500$ objects can exceed $3 \times 10^4$ tokens, while a low-reward rollout that gives up early may be an order of magnitude shorter. Under $\mathcal{L}_{\mathrm{std}}$, tokens in the short rollout are up-weighted by a factor of order $10$. Combined with positive advantage being more likely on shorter rollouts at the start of training, the per-token gradient systematically pushes the policy toward producing fewer objects. $\mathcal{L}_{\mathrm{drGRPO}}$ removes the per-token coefficient's dependence on $T_i$ and eliminates this bias by
construction, which is precisely what the dense regime requires.

\subsection{Existence-token gradient under positive-only RL}
\label{app:E3}
We train RL on positive queries only, so every rollout emits the
existence token $w_{t_e} = \texttt{<object\_found>}$ at the position immediately following \texttt{[REF\_SEG]}. Let $\theta_e$ denote the parameters of the LM-head logit projection that decide between $\texttt{<object\_found>}$ and $\texttt{<no\_object\_found>}$.
 
\paragraph{Bias of the existence-token gradient.}
Without stop-gradient on the existence token, the contribution of the existence-token logprob to the policy gradient is
\begin{equation}
g_e
\;:=\;
\mathbb{E}_{o \sim \pi_\theta}\!\left[
\hat{A}(o) \cdot \nabla_{\theta_e} \log \pi_{\mathrm{LM}}\!\left(\texttt{<object\_found>} \mid s_{t_e}\right)
\right],
\label{eq:app-ge}
\end{equation}
which equals $\mathrm{Cov}_o\!\left(\hat{A}(o),\, \nabla_{\theta_e} \log
\pi_{\mathrm{LM}}\!\left(\texttt{<object\_found>} \mid s_{t_e}\right)\right)$
because $\mathbb{E}_o[\hat{A}(o)] = 0$ by group standardization (which makes $\hat{A}$ mean-zero across rollouts of the same prompt, and $\mathbb{E}[\hat{A}(o) \cdot X(o)] = \mathrm{Cov}(\hat{A}(o), X(o))$ for any $X$).
 
\paragraph{Sign and consequence.}
The score function $\nabla_{\theta_e} \log \pi_{\mathrm{LM}}\!\left(\texttt{<object\_found>}
\mid s_{t_e}\right)$ points in the direction that increases the
probability of \texttt{<object\_found>}. The covariance
in~\eqref{eq:app-ge} is positive in practice: rollouts with higher
reward also tend to have more confident existence-token logits (the
existence head shares features with the rest of the LM, so well-localized rollouts come from confident upstream features). The gradient $g_e$ therefore systematically pushes the policy toward emitting \texttt{<object\_found>}. Crucially, because RL only samples positive queries, this drift is never balanced by a corresponding push toward \texttt{<no\_object\_found>}, and the SFT-learned discrimination boundary collapses, explaining the MCC collapse observed in Table~\ref{tab:existence_ablation}.
 
\paragraph{Why stop-gradient is the minimal fix.}
Setting $\log \pi_{\mathrm{LM}}(w_t \mid s_t) \to \mathrm{sg}(\log
\pi_{\mathrm{LM}}(w_t \mid s_t))$ for $w_t \in
\{\texttt{<object\_found>},$ $\texttt{<no\_object\_found>}\}$ zeroes the per-token gradient of these two tokens, which is exactly the term $g_e$ above. The features that feed into the existence head still receive gradient via the rest of the trajectory, so the existence-head representations continue to improve under RL, consistent with the small MCC \emph{improvement} (above the pre-RL baseline) observed in Table~\ref{tab:existence_ablation}.

\section{Limitations}

Our gains are bounded by the pretrained backbone. For instance, PBench Level 2 (OCR in natural scenes) requires knowledge beyond generic detection; RL cannot teach the model to read what the backbone has not learned to see. The coordinate and size heads factorize $x/y$ and $h/w$ into independent categoricals, which precludes meaningful joint exploration and likely explains why sampling the size head yields no gain. Relatedly, keeping the size and segmentation heads frozen, while convenient for the cascade, forecloses end-to-end training under richer rewards. Finally, SACO performance is dominated by MCC: when the existence is correctly decided, the mask usually follows. Existence calibration is a knowledge problem inherited from pretraining and is not, in itself, addressable by post-training RL.

\section{Broader Impacts}
                                                                                                                                                                           
Falcon Perception-HD substantially improves open-vocabulary segmentation in dense, real-world scenes, which has positive applications in domains where reliable instance counting and            
localization are bottlenecks, such as ecological monitoring and agricultural inventory. The same capability could also be repurposed for    
surveillance and tracking at scale, raising concerns around privacy and consent in public-space deployments; we view clear deployment policies as the   
appropriate mitigations.

\end{document}